\documentclass[pdflatex,sn-mathphys]{sn-jnl}

\jyear{2021}%

\theoremstyle{thmstyleone}%
\theoremstyle{thmstyletwo}%

\theoremstyle{thmstylethree}%

\usepackage{makecell}
\usepackage{subcaption}
\usepackage[section]{placeins}
\usepackage{tabularx}
\usepackage{multirow}
\usepackage{CJKutf8}
\usepackage{amsmath} 
\usepackage{amssymb}
\usepackage{fancyhdr}

\usepackage{xcolor}  
\usepackage{csquotes}

\usepackage{float}
\usepackage{etoolbox}
\makeatletter
\patchcmd{\ps@headings}
{\hbox to \hsize{\hfill Springer Nature 2021 \LaTeX\ template\hfill}}
{\hbox to \hsize{}}
{}
{}
\patchcmd{\ps@titlepage}
{\hbox to \hsize{\hfill Springer Nature 2021 \LaTeX\ template\hfill}}
{\hbox to \hsize{}}
{}
{}
\makeatother
\makeatletter
\patchcmd{\ps@headings}
{\hbox to \hsize{\hfill Springer Nature 2021 \LaTeX\ template\hfill}}
{\hbox to \hsize{}}
{}
{}
\patchcmd{\ps@headings}
{\hbox to \hsize{\hfill Springer Nature 2021 \LaTeX\ template\hfill}}
{\hbox to \hsize{}}
{}
{}
\patchcmd{\ps@titlepage}
{\hbox to \hsize{\hfill Springer Nature 2021 \LaTeX\ template\hfill}}
{\hbox to \hsize{}}
{}
{}

\makeatother

\usepackage{xpatch}
\makeatletter
\xpatchcmd*{\shipout}{Springer Nature 2021 \LaTeX\ template}{}{}{}
\xpatchcmd*{\@outputpage}{Springer Nature 2021 \LaTeX\ template}{}{}{}
\xpatchcmd*{\ps@headings}{Springer Nature 2021 \LaTeX\ template}{}{}{}
\xpatchcmd*{\ps@myheadings}{Springer Nature 2021 \LaTeX\ template}{}{}{}
\makeatother
\begin{document}

\begin{CJK*}{UTF8}{gbsn}

\title{Revealing the Potential of Learnable Perturbation Ensemble Forecast Model for Tropical Cyclone Prediction}

\pagestyle{plain}



\author[1]{\fnm{Jun} \sur{Liu}}\email{junliu23@m.fudan.edu.cn}
\author[2]{\fnm{Tao} \sur{Zhou}}\email{tzhou22@m.fudan.edu.cn}
\author[1]{\fnm{Jiarui} \sur{Li}}\email{jrli24@m.fudan.edu.cn}
\author[1]{\fnm{Xiaohui} \sur{Zhong}}\email{x7zhong@gmail.com}
\author[2]{\fnm{Peng} \sur{Zhang}}\email{zhang\_peng@fudan.edu.cn}
\author[2]{\fnm{Jie} \sur{Feng}}\email{fengjiefj@fudan.edu.cn}
\author[1,3]{\fnm{Lei} \sur{Chen}}\email{cltpys@163.com}
\author*[1,3]{\fnm{Hao} \sur{Li}}\email{lihao$\_$lh@fudan.edu.cn}

\affil[1]{\orgdiv{Artificial Intelligence Innovation and Incubation Institute}, \orgname{Fudan University}, \orgaddress{\city{Shanghai}, \postcode{200433}, \country{China}}}
\affil[2]{\orgdiv{Department of Atmospheric and Oceanic Sciences, Institute of
Atmospheric Sciences}, \orgname{Fudan University}, \orgaddress{\city{Shanghai}, \postcode{200438}, \country{China}}}
\affil[3]{\orgname{Shanghai Academy of Artificial Intelligence for Science}, \orgaddress{\city{Shanghai}, \postcode{200232}, \country{China}}}

\abstract{
Tropical cyclones (TCs) are highly destructive and inherently uncertain weather systems. Ensemble forecasting helps quantify these uncertainties, yet traditional systems are constrained by high computational costs and limited capability to fully represent atmospheric nonlinearity. FuXi-ENS introduces a learnable perturbation scheme for ensemble generation, representing a novel AI-based forecasting paradigm. Here, we systematically compare FuXi-ENS with ECMWF-ENS using all 90 global TCs in 2018, examining their performance in TC-related physical variables, track and intensity forecasts, and the associated dynamical and thermodynamical fields. FuXi-ENS demonstrates clear advantages in predicting TC-related physical variables, and achieves more accurate track forecasts with reduced ensemble spread, though it still underestimates intensity relative to observations. Further dynamical and thermodynamical analyses reveal that FuXi-ENS better captures large-scale circulation, with moisture turbulent energy more tightly concentrated around the TC warm core, whereas ECMWF-ENS exhibits a more dispersed distribution. These findings highlight the potential of learnable perturbations to improve TC forecasting skill and provide valuable insights for advancing AI-based ensemble prediction of extreme weather events that have significant societal impacts.
}



\maketitle

\pagestyle{plain}  


\section{Introduction}\label{sec1}

Tropical cyclones (TCs) are among the most destructive natural hazards on Earth \cite{emanuel2005increasing}, causing severe economic losses and casualties annually \cite{mendelsohn2012impact,peduzzi2012global,zhang2009tropical,doocy2013human,qin2024recent}. As highly nonlinear systems, TCs exhibit pronounced uncertainty \cite{palmer2000predicting}. While deterministic forecasts provide only single-trajectory predictions, ensemble forecasts quantify these uncertainties by sampling initial condition and model errors and accounting for day-to-day error growth variability \cite{lang2024aifs,leutbecher2008ensemble,feng2024ensemble,wang2020uncertainty,zhang2017impact}. Given the destructive impacts and intrinsic unpredictability of TCs, improving ensemble forecasting systems is critical for advancing TC prediction capabilities.

The design of the perturbation scheme is essential for ensemble forecasting 
to achieve uncertainty quantification and improve prediction skill 
\cite{buizza1999stochastic,buizza2019introduction}.
Traditional ensemble systems are built upon numerical weather prediction (NWP) 
models, where perturbation schemes are designed to represent uncertainties in 
initial conditions and physical parameterizations 
\cite{buizza1994localization,buizza1999stochastic}.
The European Centre for Medium-Range Weather Forecasts (ECMWF) ensemble prediction system (ECMWF-ENS) employs singular vectors (SV) to identify the fastest-growing perturbation modes for representing initial condition uncertainties \cite{buizza1995singular,buizza2008potential}, and incorporates the Stochastically Perturbed Parametrization Tendencies (SPPT) scheme  to characterize  model physics uncertainties \cite{wastl2019independent}.
This design yields exceptional skill in mid-latitude 500 hPa geopotential forecasts, with high anomaly correlation coefficients (ACC) in Northern Hemisphere mid-latitudes  \cite{leutbecher2008ensemble,langland2012recent,molteni1996ecmwf,palmer2019ecmwf}.
However, significant biases remain in TC forecasting \cite{aijaz2019bias, elsberry2021predicting}.
The National Centers for Environmental Prediction Global Ensemble Forecast System (NCEP-GEFS) generates initial perturbations using the ensemble Kalman filter (EnKF) approach \cite{hamill2013noaa, houtekamer2001sequential}.
GEFS performs well in short-to-medium range TC track forecasts \cite{bishop2001adaptive, zhang2009cloud}, yet systematic biases persist in TC intensity prediction \cite{zhou2019toward, hazelton2018evaluation}.
Hybrid data assimilation methods that combine Four-Dimensional Variational (4D-Var) and EnKF techniques have improved TC initialization and ensemble forecasting by leveraging flow-dependent background error covariance while maintaining the dynamical constraints of variational frameworks \cite{zhang2009coupling, zhang2012e4dvar}.
These hybrid systems exhibit enhanced performance in capturing complex TC structural evolution. 
Despite such progress \cite{buizza2005comparison}, conventional ensemble forecasting remains computationally expensive, with costs scaling linearly with ensemble size and imposing heavy burdens on high-performance computing resources  \cite{leutbecher2008ensemble,bauer2015quiet,pu2025fast}, thereby limiting its scalability, operational feasibility, and further development.  

Recent advances in machine learning have revolutionized weather forecasting, significantly alleviating the computational demands of NWP models while maintaining high accuracy.
Models such as FourCastNet \cite{pathak2022fourcastnet}, Pangu-Weather \cite{bi2022pangu}, GraphCast \cite{lam2023learning}, and FuXi \cite{chen2023fuxi} achieve deterministic forecasts with remarkable computational efficiency and skill.
However, deterministic models inherently lack the ability to quantify forecast uncertainty, limiting their use in probabilistic forecasting \cite{lang2025multi}. 
To overcome this limitation, AI-based ensemble forecasting systems have emerged. Diffusion-based GenCast \cite{price2025probabilistic} generates ensemble forecasts through multiple sampling iterations and exhibits exceptional probabilistic forecasting skill. Yet its multi-step denoising processes necessitates high computational costs relative to other machine learning models \cite{zhong2024fuxi}. Lang et al. \cite{lang2024aifs} proposed AIFS-CRPS, which incorporates the Continuous Ranked Probability Score (CRPS) as the loss function for ensemble forecasting. To mitigate bias from finite ensemble sizes during training, they introduced the almost fair CRPS. However, the reliance on limited ensemble members may still introduce sampling limitations. FuXi-ENS \cite{zhong2024fuxi} introduces a flow-dependent, learnable perturbation scheme that dynamically optimizes perturbations through data-driven learning \cite{bonavita2016evolution,houtekamer2016review}, offering a novel AI-based ensemble forecasting paradigm. Nevertheless, a systematic evaluation of this paradigm for TC prediction has not been conducted.


Here, we establish a comprehensive evaluation framework using a global dataset of 90 TCs from 2018 to systematically compare FuXi-ENS with ECMWF-ENS, the current state-of-the-art physics-based ensemble forecasting system. Our evaluation follows a progressive, multi-dimensional approach: (1) assessing forecast skill for key TC-related physical variables; (2) analyzing TC track prediction accuracy, including along-track and cross-track error evolution and the consistency between ensemble mean position errors and ensemble spread; (3) evaluating TC strike probability and intensity forecast skill; and (4) diagnosing the dynamical and thermodynamical mechanisms underlying FuXi-ENS performance through analyses of large-scale circulation evolution and perturbation energy distributions \cite{palmer1998singular,ehrendorfer1999singular}.

\section{Results}

In this study, we evaluate FuXi-ENS using 90 TCs from 2018, using ERA5 reanalysis data as initial conditions and performing twice-daily forecasts at 00 UTC and 12 UTC.
FuXi-ENS produces ensemble forecasts with 48 members, while ECMWF-ENS contains 51 members.
Both systems provide 15-day forecasts at 0.25° spatial resolution with 6-hour intervals, covering all 90 global TCs in 2018.
A comprehensive, multi-dimensional evaluation is conducted, including analyses 
of TC-related physical variables, ensemble track forecast errors, TC strike probability and intensity forecast skill, and the underlying dynamical and 
thermodynamical fields. In this evaluation, ERA5 reanalysis data and the 
International Best Track Archive for Climate Stewardship (IBTrACS) records 
serve as ground truth for meteorological field and track validation, respectively.

\subsection{TC-Related Physical Variables Performance}
To comprehensively evaluate FuXi-ENS performance in TC prediction, we first assess its forecast skill for key physical variables closely related to TC behavior. We examine six variables spanning surface and upper-level processes: mean sea level pressure (MSL), 2m temperature (T2M), 10m wind speed (WS10M), 500 hPa geopotential (Z500), 850 hPa temperature (T850), and 850 hPa wind speed (WS850). These variables capture critical aspects of TC dynamics: MSL reflects the pressure gradient driving TC intensity; T2M and T850 characterize the thermal structure essential for TC development; WS10M and WS850 represent wind circulation patterns governing TC organization; and Z500 captures the large-scale flow patterns influencing TC motion \cite{gray1968global, emanuel1986air, done2022response}. To quantify probabilistic forecast skill, we employ the Receiver Operating Characteristic Area Skill Score (ROCASS), which measures a model's ability to discriminate extreme events \cite{richardson2000skill}. Scores are computed across all lead times (6–360 h) for all 730 initialization times in 2018 (00 and 12 UTC daily), separately for upper and lower terciles of each variable.


Figs. 1 and 2 present ROCASS for the upper and lower terciles of surface and upper-level variables, respectively. For both FuXi-ENS and ECMWF-ENS, ROCASS declines progressively with increasing forecast lead time, reflecting the expected deterioration of forecast skill across all variables examined. This degradation is steepest for wind fields (WS10M, WS850), whose ROCASS values fall below 0.2 by day 15 (Fig. 1e, f; Fig. 2e, f), indicating greater uncertainty and relatively lower predictability. In contrast, mean sea level pressure (MSL), temperature (T2M, T850), and geopotential fields (Z500) maintain higher skill, with ROCASS values remaining near 0.4 at day 15 (Fig. 1a–d; Fig. 2a–d). FuXi-ENS demonstrates marginally higher ROCASS values for wind speed variables (WS10M, WS850) compared to ECMWF-ENS, with the difference being more pronounced for both upper and lower extreme terciles, indicating superior ability to discriminate both strong and weak extreme wind events. For T2M tercile events, however, FuXi-ENS outperforms ECMWF-ENS through day 8 but exhibits slightly lower skill beyond day 9.

\begin{figure}[H]
    \centering
    \includegraphics[width=\linewidth, height=0.555\linewidth]{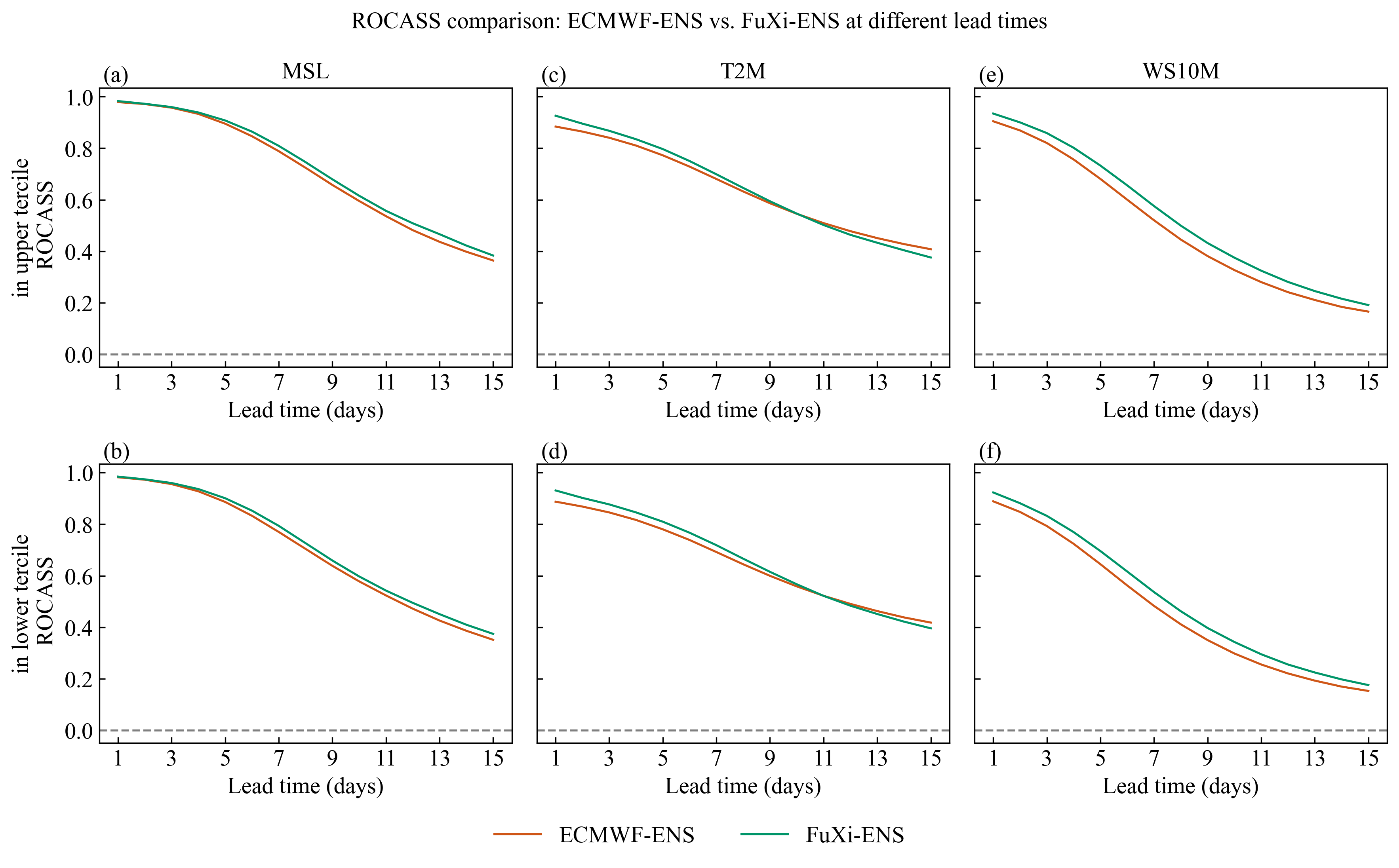}
    \caption{\textbf{ROCASS comparison between FuXi-ENS (green) and ECMWF-ENS (orange) for surface variables over 15-day forecasts with 6-hourly evaluation intervals.} ROCASS are evaluated for the upper tercile (top panel) and lower tercile (bottom panel) of three surface variables: mean sea level pressure (MSL, left panel), 2m temperature (T2M, middle panel), and 10m wind speed (WS10M, right panel).}
    \label{model}    
\end{figure}
\vspace{-3em}


\begin{figure}[H]
    \centering
    \includegraphics[width=\linewidth, height=0.555\linewidth]{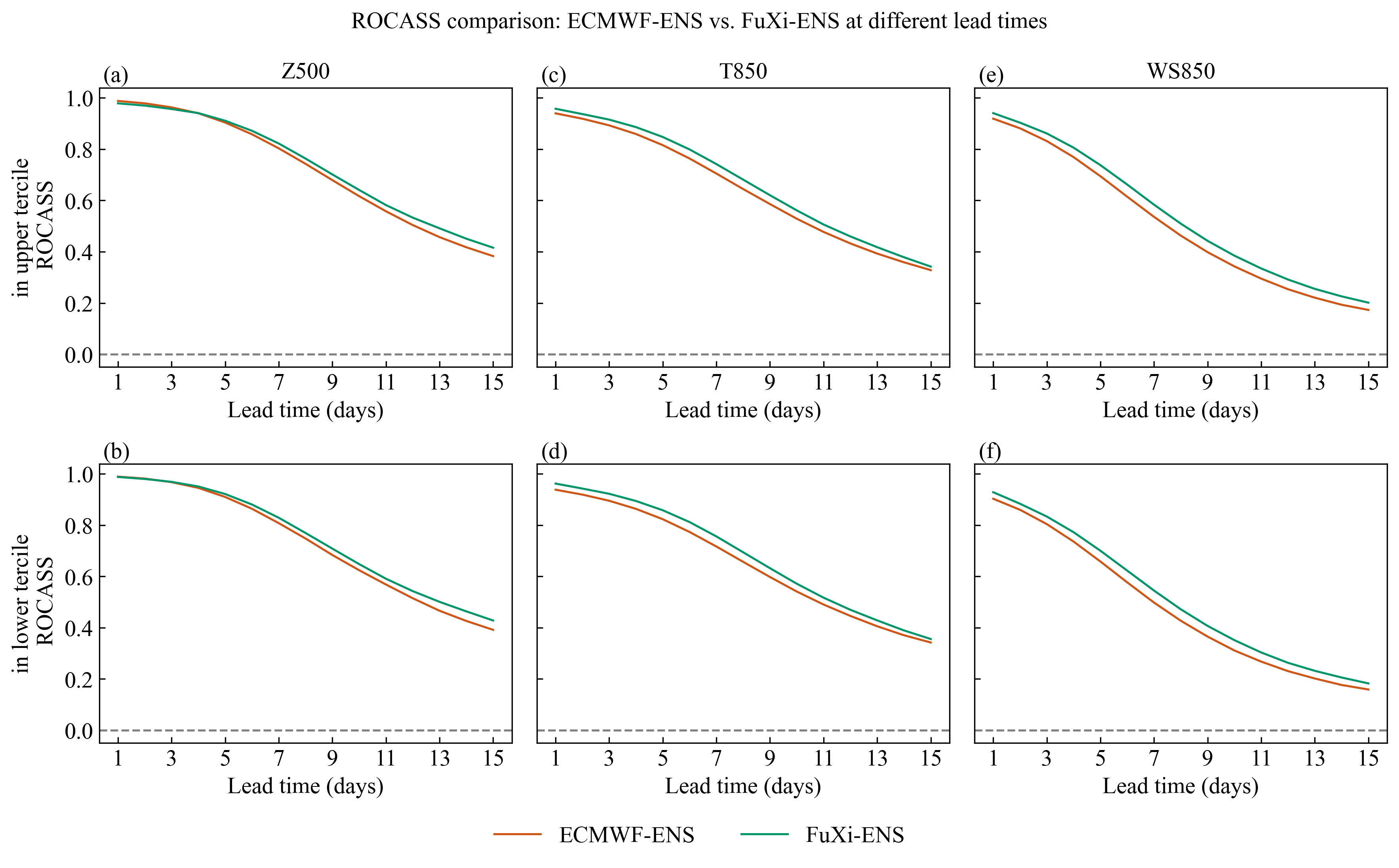}
    \caption{\textbf{ROCASS comparison between FuXi-ENS (green) and ECMWF-ENS (orange) for upper-level variables over 15-day forecasts with 6-hourly evaluation intervals.
} ROCASS are evaluated for the upper tercile (top panel) and the lower tercile (bottom panel) of three upper-level atmospheric variables: 500 hPa geopotential (Z500, left panel), 850 hPa temperature (T850, middle panel), and 850 hPa wind speed (WS850, right panel).}
    \label{model}    
\end{figure}

\subsection{Ensemble Forecast Track Error Analysis for TCs}

Building upon the comprehensive ROCASS evaluation that demonstrated FuXi-ENS's superior performance in TC-related physical variables, we now present a detailed track error analysis for TC ensemble forecasts. Our TC tracking algorithm (Supplementary Fig. 10) identifies cyclone centers based on minimum MSL and maximum relative vorticity at 850 hPa. We apply this algorithm to all 90 TCs in 2018, tracking FuXi-ENS prediction fields from initialization throughout their life cycles (00 and 12 UTC daily), yielding track forecasts for each ensemble member. Accumulated ensemble mean position error ($\mathrm{AccError}_{\mathrm{TC}}$) quantifies the deviation between ensemble mean forecasts and IBTrACS observations, while accumulated ensemble spread ($\mathrm{AccSpread}_{\mathrm{TC}}$) measures the dispersion of individual ensemble members relative to the ensemble mean. To elucidate directional biases, we further decompose ensemble mean track errors into along-track error (AT, the component parallel to the observed track direction) and cross-track error (CT, the component perpendicular to the observed track direction), as illustrated in Supplementary Fig. 11.

Fig. 3 reveals the temporal evolution of four metrics across forecast lead times: $\mathrm{AccError}_{\mathrm{TC}}$, $\mathrm{AccSpread}_{\mathrm{TC}}$, AT, and CT. While both systems exhibit increasing $\mathrm{AccError}_{\mathrm{TC}}$ with lead time, FuXi-ENS consistently achieves significantly lower median errors and markedly tighter error distributions than ECMWF-ENS (Fig. 3a). $\mathrm{AccSpread}_{\mathrm{TC}}$ similarly increases with lead time, yet FuXi-ENS maintains considerably smaller spread than ECMWF-ENS throughout all forecast horizons (Fig. 3b). Directional error decomposition provides further insights into forecast biases: AT quantifies TC positional deviation parallel to the observed motion, with positive values indicating forecasts ahead of observations and negative values indicating lag. Both systems exhibit systematic lag bias (Fig. 3c), but ECMWF-ENS shows substantially greater median lag and broader error distributions. In contrast, CT quantifies deviation perpendicular to the observed motion, exhibits near-zero medians for both systems, indicating negligible systematic lateral bias, though ECMWF-ENS displays considerably wider distributions (Fig. 3d), reflecting greater forecast uncertainty in the lateral dimension.

\begin{figure}[H]
    \centering
    \includegraphics[width=\linewidth]{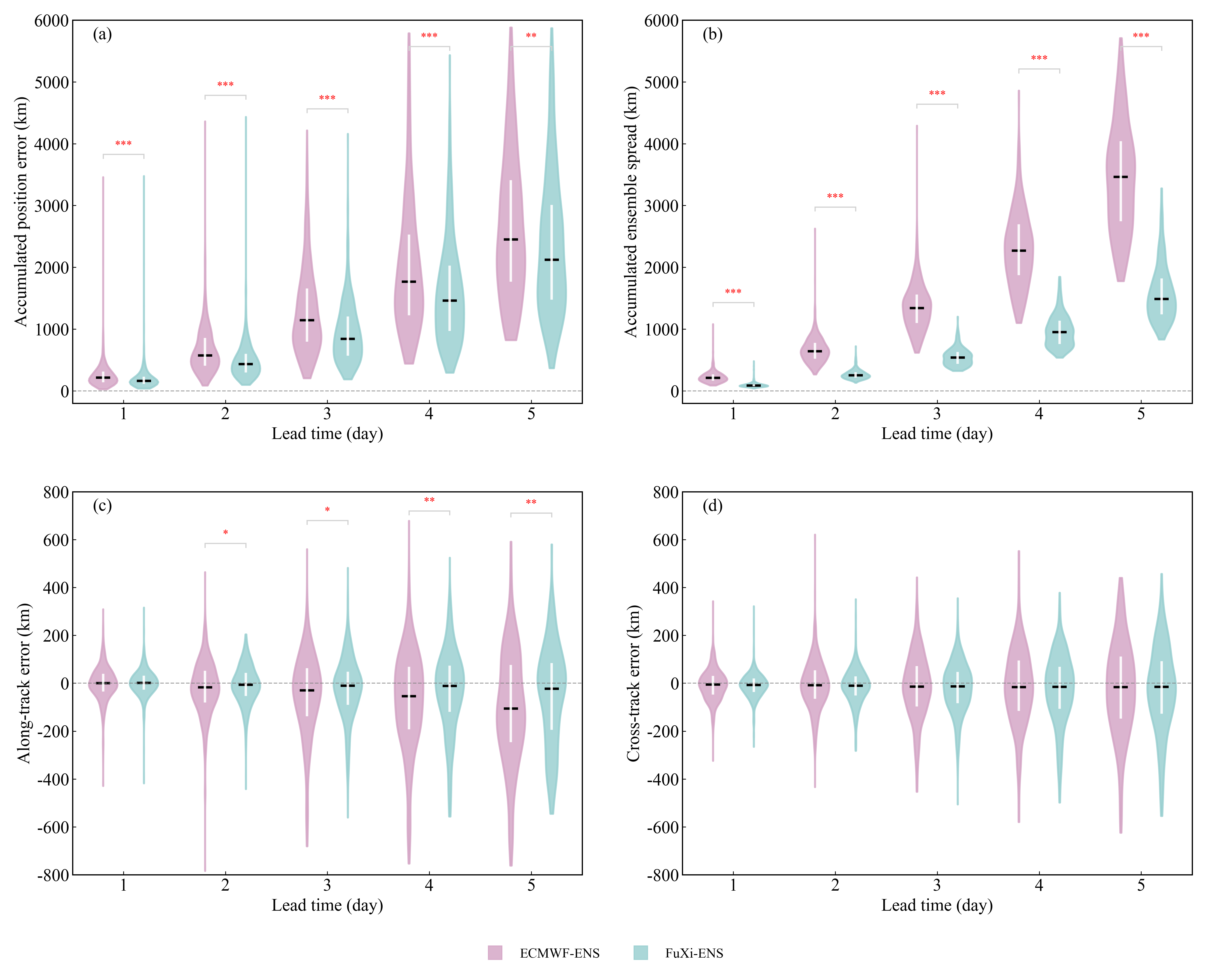}
    \caption{\textbf{Violin plots of TC track forecast errors in 5-day forecasts for ECMWF-ENS (purple) and FuXi-ENS (teal).}
   (\textbf{a}), accumulated ensemble mean position error ($\mathrm{AccError}_{\mathrm{TC}}$, km). (\textbf{b}), accumulated ensemble spread ($\mathrm{AccSpread}_{\mathrm{TC}}$, km). (\textbf{c}), along-track error (AT, km). (\textbf{d}), cross-track error (CT, km). Violin plots show the distribution characteristics of each error metric across forecast lead times, with dashed horizontal lines indicating the median and interquartile ranges marked. Purple represents ECMWF-ENS forecasts, and teal represents FuXi-ENS forecasts. Asterisks indicate statistical significance levels ($*$ $\mathrm{p} < 0.05$, $**$ $\mathrm{p} < 0.01$, $***$ $\mathrm{p} < 0.001$) based on Mann-Whitney U tests comparing the two ensemble systems.}
    \label{model}    
\end{figure}

Fig. 4 further compares the relationships between $\mathrm{AccError}_{\mathrm{TC}}$ and $\mathrm{AccSpread}_{\mathrm{TC}}$, as well as between AT and CT for both systems. FuXi-ENS exhibits more concentrated scatter points, with most samples showing $\mathrm{AccError}_{\mathrm{TC}}$ exceeding $\mathrm{AccSpread}_{\mathrm{TC}}$ (Fig. 4a). In contrast, ECMWF-ENS displays wider scatter distribution, with most samples showing $\mathrm{AccSpread}_{\mathrm{TC}}$ exceeding $\mathrm{AccError}_{\mathrm{TC}}$ (Fig. 4b). For the relationship between AT and CT, scatter points closer to the origin indicate smaller systematic bias. FuXi-ENS shows more concentrated distribution near the origin with relatively uniform scatter across four quadrants, indicating smaller biases in both directions (Fig. 4c). In contrast, ECMWF-ENS exhibits more dispersed scatter points with significantly more samples in the negative AT direction, revealing obvious lagging bias. Additionally, ECMWF-ENS shows more outliers, further reflecting forecast instability (Fig. 4d). In summary, FuXi-ENS outperforms ECMWF-ENS in TC track simulation with smaller accumulated ensemble mean position errors, more concentrated ensemble forecasts, and more reasonable along-track error characteristics.

\begin{figure}[H]
    \centering
    \includegraphics[width=0.78\linewidth]{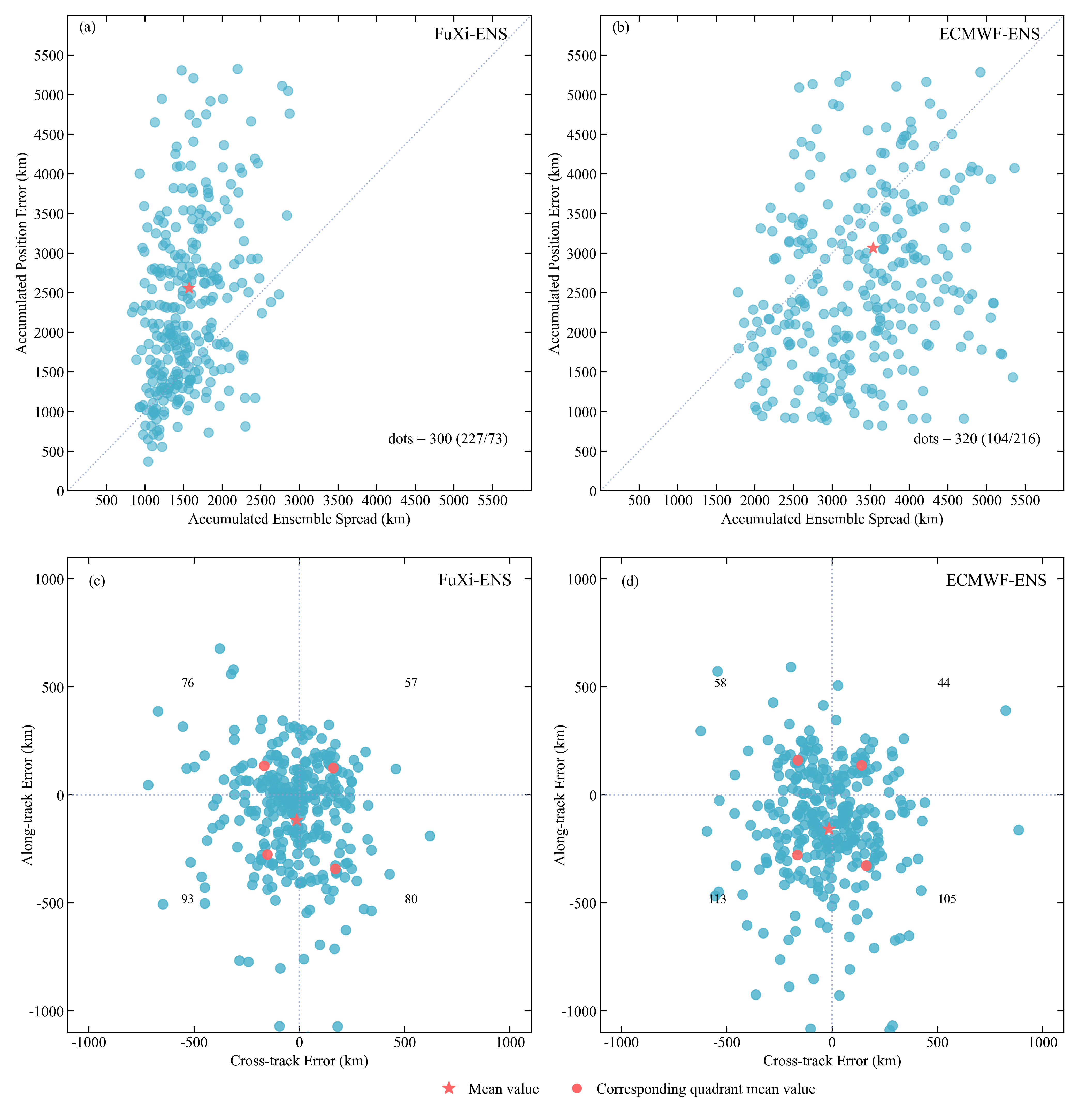}
    \caption{\textbf{Scatter plots comparing $\mathrm{\textbf{AccError}}_{\mathrm{\textbf{TC}}}$ against $\mathrm{\textbf{AccSpread}}_{\mathrm{\textbf{TC}}}$, and AT against CT at a 5-day forecast lead time for FuXi-ENS and ECMWF-ENS.} Scatter plot distribution of $\mathrm{AccError}_{\mathrm{TC}}$ versus $\mathrm{AccSpread}_{\mathrm{TC}}$ for (\textbf{a}) FuXi-ENS and (\textbf{b}) ECMWF-ENS, and AT versus CT for (\textbf{c}) FuXi-ENS and (\textbf{d}) ECMWF-ENS. Each dot represents an individual TC forecast initialization. Pentagrams indicate overall means; in upper panels, dashed lines show 1:1 references with point counts for upper/lower regions; in lower panels, solid circles show quadrant means, numbers indicate counts per quadrant, and dashed lines show zero-error references.}
    \label{model}    
\end{figure}

\subsection{TC strike probability and intensity forecast skill}
Given FuXi-ENS's superior overall performance in TC-related physical variables and track error analysis, we further examine how these improvements manifest in specific individual TC cases. Taking typhoon KONG-REY as an example, this typhoon formed in the central Northwest Pacific in late September, initially following a typical westward track, gradually moving northwestward and approaching east of the Philippines. From September 30 to October 2, it gradually turned northward; around October 4, it further shifted to a northeastward direction and finally approached the southern part of the Korean Peninsula on October 6 before dissipating. The evolution of this TC was influenced by the complex interaction between the subtropical high and mid-latitude trough systems, resulting in complex track changes but with typical representativeness. 


The ensemble forecast results demonstrate that FuXi-ENS closely reproduces the best track (IBTrACS), accurately capturing the complete evolution sequence of westward movement, northward recurvature, and northeastward turning, particularly during the critical turning phase (October 2–4). Notably, the ensemble probability distribution remains relatively concentrated (Fig. 5a). In contrast, ECMWF-ENS exhibits larger simulation deviations, with its ensemble mean track showing a pronounced westward bias during the initial phase. Substantial deviations also occur during the recurvature phase, failing to accurately represent the typhoon's northeastward turning position and recurvature magnitude. Furthermore, ECMWF-ENS displays a significantly wider ensemble probability distribution than FuXi-ENS, with particularly strong track dispersion following northward recurvature (Fig. 5b). This comparative analysis of Typhoon KONG-REY demonstrates that FuXi-ENS significantly outperforms ECMWF-ENS in both track simulation accuracy and ensemble probability distribution concentration. This superior performance is further corroborated by comparative analyses of representative cases, including Hurricane Florence (Atlantic Basin) and Hurricane Bud (Eastern Pacific Basin) (Supplementary Figs. 1, 2), highlighting the model's robust generalization capability across different oceanic basins.

\begin{figure}[H]
    \centering
    \includegraphics[width=\linewidth]{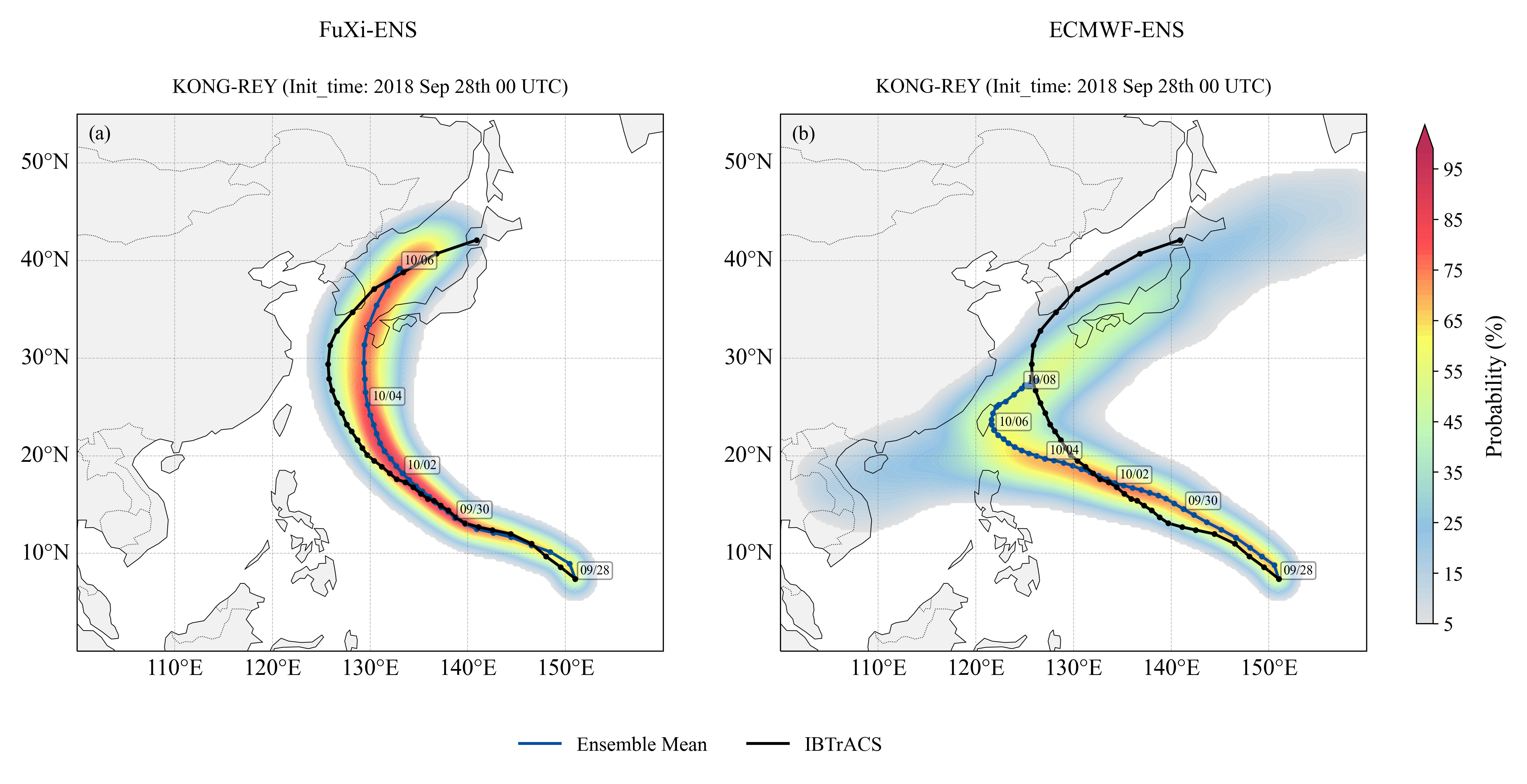}
    \caption{\textbf{TC strike probability forecast comparison for Typhoon KONG-REY.} Probabilistic distribution of TC strike probability for Typhoon KONG-REY based on ensemble forecasts initialized at 00 UTC on September 28, 2018 from (\textbf{a}) FuXi-ENS and (\textbf{b}) ECMWF-ENS. Blue lines denote ensemble mean tracks; black lines indicate  best track observations (IBTrACS). Shading indicates TC strike probability $(\mathrm{\%})$. Date labels mark the temporal evolution of the track.}
\label{fig:kongrey_strike_probability}
  
\end{figure}

Beyond TC track forecasting and strike probability prediction, accurate intensity prediction is equally critical. Statistical verification across all cases reveals that FuXi-ENS systematically underestimates TC intensity relative to both ECMWF-ENS and best track observations (IBTrACS), reflecting a persistent challenge in current AI-based weather prediction models \cite{zhong2024fuxi, dulac2024assessing, liu2024evaluation}. To systematically evaluate the ensemble forecasting capability of FuXi-ENS for TCs, we present a representative case examining intensity evolution. ERA5 reanalysis data are used as the verification benchmark to ensure consistency with subsequent circulation analyses. Fig. 6 illustrates the intensity evolution of Typhoon KONG-REY, with corresponding results for Typhoon TRAMI presented in Supplementary Fig. 3. For the 10m maximum wind speed evolution, FuXi-ENS closely tracks ERA5 throughout the development phase (days 0–4) and accurately captures the timing of peak intensity during days 4–5, while maintaining a relatively narrow ensemble spread. In contrast, ECMWF-ENS exhibits considerably greater ensemble dispersion (Fig. 6a, b). The minimum central pressure evolution displays similar characteristics. FuXi-ENS reproduces the pressure deepening process closely matching ERA5 during days 0–4 and accurately captures the timing of minimum pressure, while maintaining a concentrated ensemble distribution. Conversely, ECMWF-ENS shows substantially greater ensemble dispersion in pressure evolution, indicating higher inter-member uncertainty (Fig. 6c, d).

\begin{figure}[H]
    \centering
    \includegraphics[width=\linewidth]{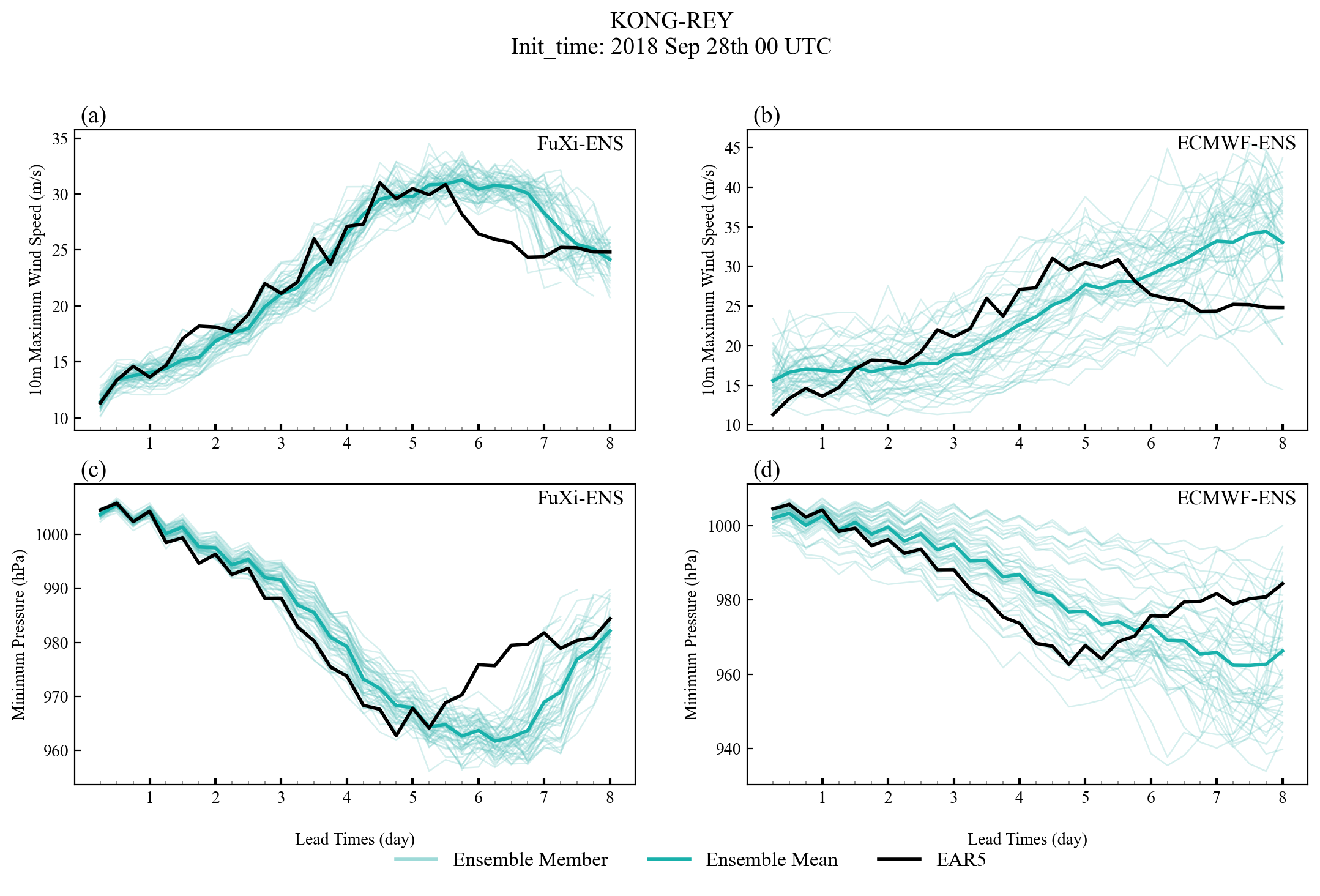}
    \caption{\textbf{Intensity forecast comparison for typhoon KONG-REY.}
    10m maximum wind speed variation with forecast lead time for (\textbf{a}) FuXi-ENS and (\textbf{b}) ECMWF-ENS initialized at 00 UTC on September 28, 2018. Minimum pressure variation with forecast lead time for (\textbf{c}) FuXi-ENS and (\textbf{d}) ECMWF-ENS. Light green thin lines represent ensemble member forecasts, dark green thick lines represent ensemble mean forecasts, and black thick lines represent ERA5 reanalysis forecasts.}
    \label{model} 
\end{figure}
\subsection{TC-Related Dynamical Analysis}


Based on the preceding analyses, FuXi-ENS outperforms ECMWF-ENS across both statistical metrics and case studies, more accurately reproducing the evolution of TC tracks and intensities. Since TC development is governed by complex dynamical and thermodynamical interactions, where large-scale steering flows primarily determine TC trajectories and sufficient moisture supports maintenance and intensification, we further examine whether FuXi-ENS more realistically simulates the associated atmospheric circulation and moisture distributions. This analysis elucidates the physical mechanisms underlying its superior performance.



Using KONG-REY as an example, FuXi-ENS more faithfully reproduces its \enquote{westward–northward–northeastward} track pattern compared to ECMWF-ENS. To diagnose the atmospheric circulation features responsible for these differences, we analyze the evolution of wind and geopotential height fields at multiple levels. Fig. 7 shows the 500 hPa wind speed (shading), geopotential height (contours), and horizontal wind vectors for KONG-REY initialized at 00 UTC on September 28, 2018, at lead times of 1, 3, 5, and 7 days from FuXi-ENS, ECMWF-ENS, and ERA5. Corresponding 850 hPa and surface level results are presented in Supplementary Figs. 4, 5. The 850 hPa circulation evolution for TRAMI case exhibits similar patterns (Supplementary Fig. 6).

At 500 hPa, the Western Pacific Subtropical High (WPSH) is the dominant system governing TC motion. Before October 5, both FuXi-ENS and ECMWF-ENS capture a reasonably strong and well-positioned WPSH ridge that steers KONG-REY northwestward, though FuXi-ENS more accurately represents the WPSH's spatial extent compared to ERA5 (Fig. 7a-c, e-g, i-k). On October 5, both ERA5 reanalysis and FuXi-ENS show a deep trough over Japan and Korea, forming a \enquote{high--low cooperation} pattern with the WPSH that creates favorable steering flow for KONG-REY's northward and subsequent eastward recurvature (Fig. 7d, l). In contrast, ECMWF-ENS fails to reproduce this trough, resulting in a relatively zonal circulation pattern that lacks the necessary recurvature mechanism (Fig. 7h).

Similar behavior is observed at 850 hPa. Prior to October 5, both systems reasonably simulate the WPSH's position and the associated northwestward steering flow for KONG-REY. By October 5, FuXi-ENS maintains the curvature and position of the 1520 gpm geopotential height contour consistent with ERA5, whereas ECMWF-ENS produces a relatively zonal contour that inhibits the TC's northeastward recurvature (Supplementary Fig. 4). At the surface, both ERA5 and FuXi-ENS exhibit pronounced curvature in the 1016 hPa isobar, reflecting the WPSH's modulation of TC motion. In contrast, ECMWF-ENS displays relatively straight isobars, inadequately capturing the WPSH's spatial structure and underestimating its intensity (Supplementary Fig. 5).

Likewise, FuXi-ENS also more accurately captures the large-scale circulation associated with Typhoon TRAMI (Supplementary Fig. 6).
Overall, the dynamical analyses demonstrate that FuXi-ENS reproduces both TC tracks and the associated steering flows with higher fidelity, whereas ECMWF-ENS exhibits systematic biases in simulating key circulation systems, which may lead to inferior track forecasts.

\begin{figure}[H]
    \centering
    \includegraphics[width=\linewidth]{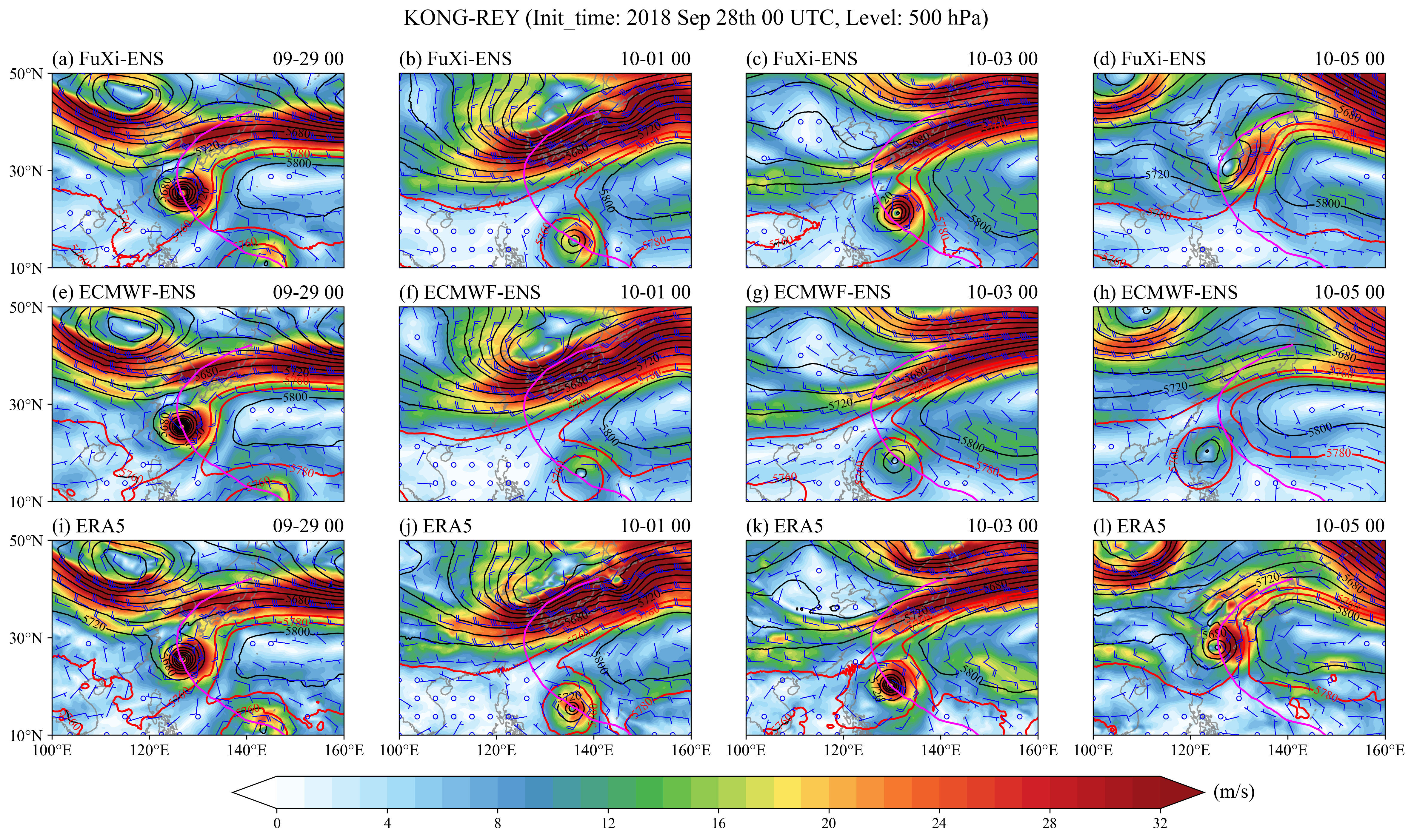}
    \caption{\textbf{Evolution of 500 hPa atmospheric circulation patterns for typhoon KONG-REY with forecast lead time.}
    Evolution of KONG-REY's 500 hPa wind speed (shading, m/s), geopotential height (contours, gpm, 40 gpm intervals), and horizontal wind (vectors, m/s) initialized at 00 UTC on September 28, 2018, for September 29 (first column), October 1 (second column), October 3 (third column), and October 5 (fourth column) from FuXi-ENS (first row), ECMWF-ENS (second row), and ERA5 (third row). Red contours indicate the 5760 and 5780 gpm contour line, and magenta lines represent best tracks (IBTrACS).}
    \label{model}    
\end{figure}


We further compare the ensemble mean and individual member simulations of specific geopotential height contour evolution for KONG-REY with consistent initialization time and lead times, as shown in Fig. 8. Similar results for TRAMI are shown in Supplementary Fig. 7.

Although both ensembles show increasing spread with lead time, FuXi-ENS members remain more tightly clustered, indicating greater internal consistency (Fig. 8a–d). In contrast, ECMWF-ENS exhibits substantially greater dispersion, with some members showing considerable deviations that reduce the representativeness of the ensemble mean (Fig. 8e–h). These results demonstrate that FuXi-ENS not only achieves higher accuracy but also maintains superior coherence among ensemble members.

\begin{figure}[H]
    \centering
    \includegraphics[width=\linewidth]{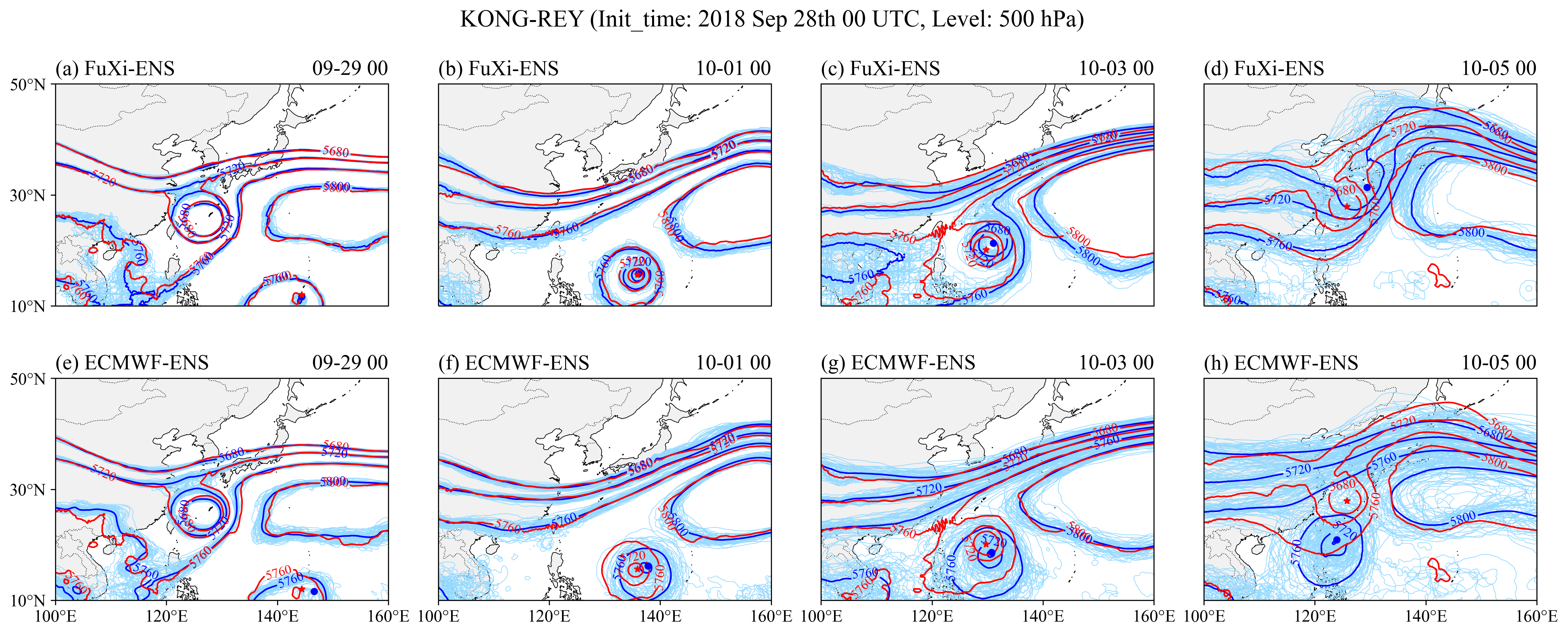}
    \caption{\textbf{Evolution of 500 hPa geopotential height ensemble forecasts for typhoon KONG-REY with forecast lead time.}
    Evolution of KONG-REY's 500 hPa geopotential height (contours, gpm) initialized at 00 UTC on September 28, 2018, for September 29 (first column), October 1 (second column), October 3 (third column), and October 5 (fourth column) from FuXi-ENS (first row), ECMWF-ENS (second row). Light blue lines represent individual ensemble members, dark blue lines represent the ensemble mean, and red lines represent ERA5 reanalysis. Four contour lines are shown: 5680 gpm, 5720 gpm, 5760 gpm and 5800 gpm. Blue dots denote the ensemble mean track and red stars denote the best track (IBTrACS).}
    \label{model}    
\end{figure}

\subsection{TC-Related Thermodynamical Analysis}

The evolution of TCs depends not only on accurate representation of dynamical fields but also critically on the evolution of thermal fields. Moisture turbulent energy (MTE), a key diagnostic of atmospheric thermal and moisture states, effectively reflects the energy distribution related to TCs.

Fig. 9 shows the evolution of 850 hPa MTE (shading), temperature (contours), and horizontal wind (vectors) for KONG-REY from FuXi-ENS and ECMWF-ENS with the same initialization time and lead times. Initially, at lead times of 1–3 days (September 29–October 1), both FuXi-ENS and ECMWF-ENS capture enhanced MTE near the TC warm core. However, ECMWF-ENS also produces extensive high MTE values across the mid- to high-latitude regions, while FuXi-ENS exhibits more compact MTE concentrated around the TC core with weaker peripheral perturbations (Fig. 9a, b, e, f).
As the forecast lead time extends, the TC moves northward and begins to interacts with mid-latitude frontal zones and jet streams.
During this transition, at a lead time of 5 days (October 3), ECMWF-ENS generates stronger and more continuous high-MTE bands, indicating greater sensitivity to frontal perturbations and higher uncertainty, whereas FuXi-ENS maintains weaker and more localized MTE (Fig. 9c, g).
In the later forecast period, at a lead time of 7 days (October 5), as the TC structure gradually dissipates, FuXi-ENS exhibits significantly enhanced MTE with perturbation energy mainly concentrated in the mid-high latitude jet region, while ECMWF-ENS continues to maintain high MTE across both frontal and jet regions (Fig. 9d, h).
Similar patterns are observed for FLORENCE and TRAMI (Supplementary Figs. 8 and 9).
Overall, FuXi-ENS concentrates MTE evolution within the TC core, while ECMWF-ENS shows broader perturbation diffusion across larger spatial scales, which may contribute to its larger ensemble spread.
These findings provide valuable guidance for refining ensemble perturbation schemes.

\begin{figure}[H]
    \centering
    \includegraphics[width=\linewidth]{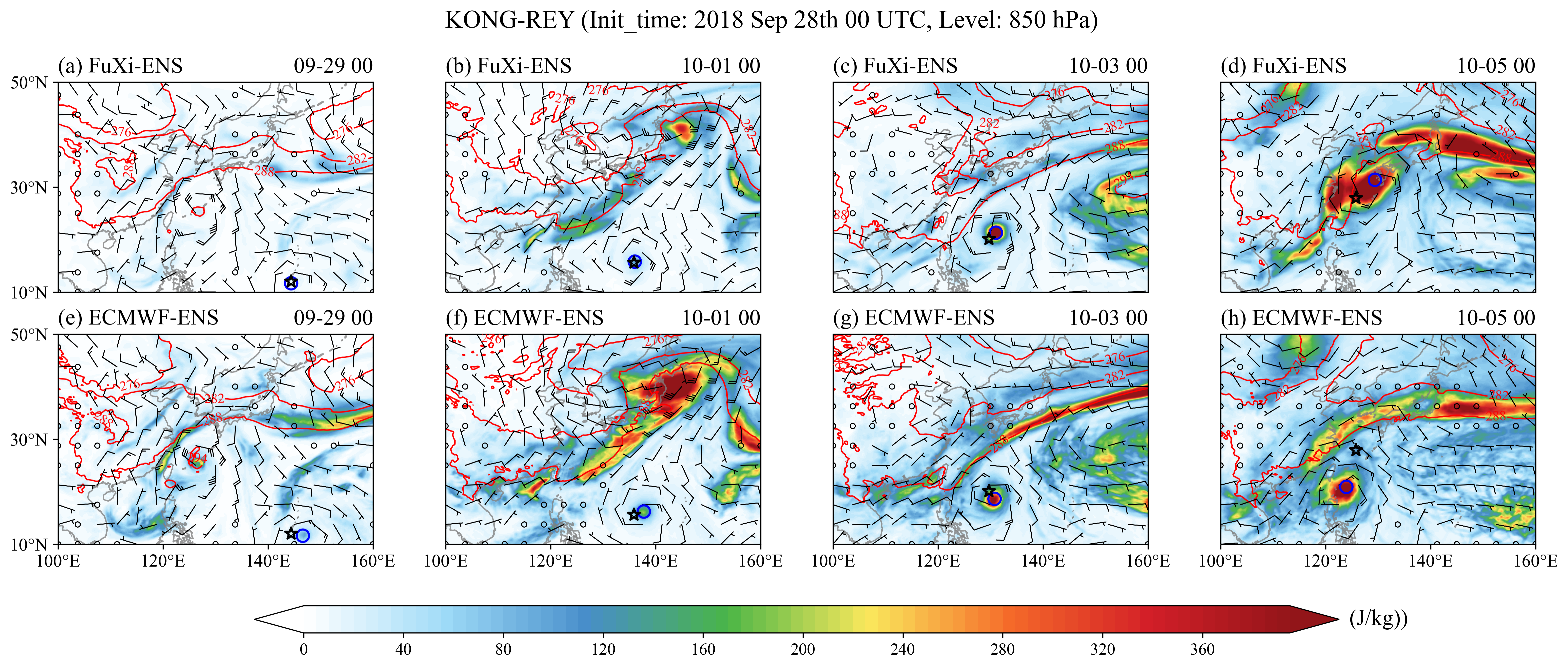}
    \caption{\textbf{Evolution of 850 hPa MTE for typhoon KONG-REY with forecast lead time.}
    Evolution of KONG-REY's 850 hPa MTE (shading, J/kg), temperature (contours, K), and horizontal wind (vectors, m/s) initialized at 00 UTC on September 28, 2018, for September 29 (first column), October 1 (second column), October 3 (third column), and October 5 (fourth column) from FuXi-ENS (first row), ECMWF-ENS (second row). Blue circles denote the ensemble mean track and black stars denote the best track (IBTrACS).}
    \label{model}    
\end{figure}

\section{Discussion}

Systematic evaluation of FuXi-ENS's learnable perturbation generation scheme demonstrates significant advantages across multiple dimensions of TC forecasting. For TC-related physical variables, FuXi-ENS generally achieves superior performance relative to ECMWF-ENS. Track forecast analysis reveals that FuXi-ENS exhibits substantially lower accumulated ensemble mean position errors and reduced ensemble spread compared to ECMWF-ENS. For intensity forecasts, FuXi-ENS reproduces TC intensity evolution more faithfully than ECMWF-ENS when evaluated against ERA5 reanalysis data. Dynamic and thermodynamic analyses further elucidate the physical mechanisms underlying FuXi-ENS's superior performance. Using Typhoon KONG-REY as a case study, FuXi-ENS more accurately captures the evolution of steering flows associated with the TC, including interactions between the subtropical high and mid-latitude troughs. Additionally, detailed analysis of MTE distribution reveals distinct characteristics between the two ensemble systems: FuXi-ENS exhibits highly concentrated MTE within the TC warm-core region, whereas ECMWF-ENS demonstrates broader spatial MTE distribution. These differences may directly explain the smaller ensemble spread in FuXi-ENS, providing valuable insights for optimizing ensemble perturbation schemes in future operational systems.

The superior track forecasts but relatively weaker intensity predictions of FuXi-ENS reflect a common trade-off in current machine learning-based forecasting systems. The improved track accuracy is primarily attributed to FuXi-ENS's ability to capture large-scale steering flow patterns through flow-dependent learnable perturbations, with this advantage being most pronounced at short-to-medium forecast lead times.
In contrast, the limited skill in intensity prediction can be attributed to two fundamental challenges. First, systematic underestimation of TC intensities in ERA5 relative to observational benchmarks introduces biased training targets that fundamentally limit model capability.
Second, the prevalent use of L1 or L2 loss functions introduces a \enquote{double penalty} problem \cite{subich2025fixing}, whereby spatially smoothed predictions are systematically favored over sharp intensity gradients. Future improvements may benefit from three key developments: (1) loss functions that mitigate the double penalty problem while preserving sensitivity to intensity gradients; (2) incorporation of higher-resolution datasets to better resolve extreme events; and (3) end-to-end optimization strategies specifically tailored for extreme event prediction. This approach would require leveraging high-quality, high-resolution datasets (such as high-resolution extreme event datasets) as training targets, with optimization strategies tailored to specific physical processes. Such targeted training may enable more precise characterization of atmospheric circulation patterns and improved capture of fine-scale features critical to extreme events.  This paradigm shift may ensure that models better understand and accurately predict extreme weather phenomena that have significant societal impacts.

Overall, this study demonstrates that machine learning-based learnable perturbation techniques represent a significant advancement in TC ensemble forecasting. Beyond the demonstrated improvements in track prediction accuracy, FuXi-ENS exhibits enhanced capability in representing the complex dynamical and thermodynamical structures of TCs. These findings suggest that learnable perturbation schemes offer a promising pathway toward more reliable operational forecasting systems, with potential applications to high-impact extreme weather events that require ensemble uncertainty quantification.

\section{Methods} 

\subsection{Data} 

This study establishes a comprehensive evaluation framework to systematically assess the performance of the FuXi-ENS learnable perturbation-based ensemble forecasting system for tropical cyclone prediction. The framework integrates initial condition perturbations, benchmark model comparisons, and observational verification, providing a scientific foundation for objectively evaluating flow-dependent perturbation schemes.

As a deep learning-based ensemble system, FuXi-ENS depends critically on high-quality initial conditions. Here, we adopt the fifth generation of the European Centre for Medium-Range Weather Forecasts (ECMWF) reanalysis dataset (ERA5) \cite{hersbach2020era5} as model input. ERA5 provides atmospheric fields at $0.25^{\circ}$ spatial resolution with 6-hourly temporal resolution, delivering accurate atmospheric states for machine learning applications. Initialized with ERA5 fields, FuXi-ENS generates 15-day ensemble forecasts at 6-hour intervals.

For benchmarking, we select the ECMWF ensemble prediction system (ECMWF-ENS) as a reference. ECMWF-ENS has been extensively validated in previous studies and is widely regarded as one of the most authoritative operational ensemble forecasting systems for tropical cyclone track prediction. Historical ECMWF-ENS forecasts are obtained from the WeatherBench2 platform (\url{gs://weatherbench2/datasets/ifs_ens/2018-2022-1440x721.zarr}), covering the period 2018–2022 at $1440 \times 721$ spatial resolution with 6-hourly output intervals.

Tropical cyclone track and intensity predictions serve as core components of forecast evaluation. For FuXi-ENS, a tracking algorithm is employed to extract TC tracks and intensities, with detailed descriptions provided in Section 4.3. Both ECMWF high-resolution deterministic forecasts (ECMWF-HRES) and ensemble forecasts (ECMWF-ENS) TC track data are retrieved from the TIGGE (THORPEX Interactive Grand Global Ensemble) archive in XML format (\url{https://confluence.ecmwf.int/display/TIGGE/Tools}).

For observational verification, we use the International Best Track Archive for Climate Stewardship (IBTrACS) as ground truth (\url{https://www.ncei.noaa.gov/products/international-best-track-archive}). Maintained by the World Meteorological Organization, IBTrACS compiles official best track records of global tropical cyclones, including 3-hourly center locations, maximum sustained winds, and minimum central pressures \cite{knapp2010international,kenneth2019international}. The dataset undergoes rigorous quality control and multi-agency cross-validation, establishing it as the authoritative reference for TC research. Comparative analysis against IBTrACS enables objective assessment of track accuracy, intensity prediction skill, and probabilistic forecast performance, providing robust evidence for evaluating learnable perturbation techniques in extreme weather forecasting.
\subsection{FuXi-ENS model} \label{fuxi.arch}

This study employs the FuXi-ENS model to explore its performance in TC track prediction. The overall framework and evaluation methodology are illustrated in Fig. 10. FuXi-ENS leverages machine learning to address ensemble forecasting challenges, offering fast and efficient probabilistic predictions through learnable perturbations that capture flow-dependent characteristics. Previous studies have demonstrated its superior performance relative to ECMWF-ENS across multiple probabilistic metrics, including continuous ranked probability score (CRPS) and Brier score \cite{zhong2024fuxi}. In this work, we comprehensively assess FuXi-ENS's forecasting skill for tropical cyclones using a dataset of 90 TCs from 2018. The technical architecture and key innovations of FuXi-ENS are described below.

\begin{figure}[!htbp]
    \centering
    \includegraphics[width=\linewidth]{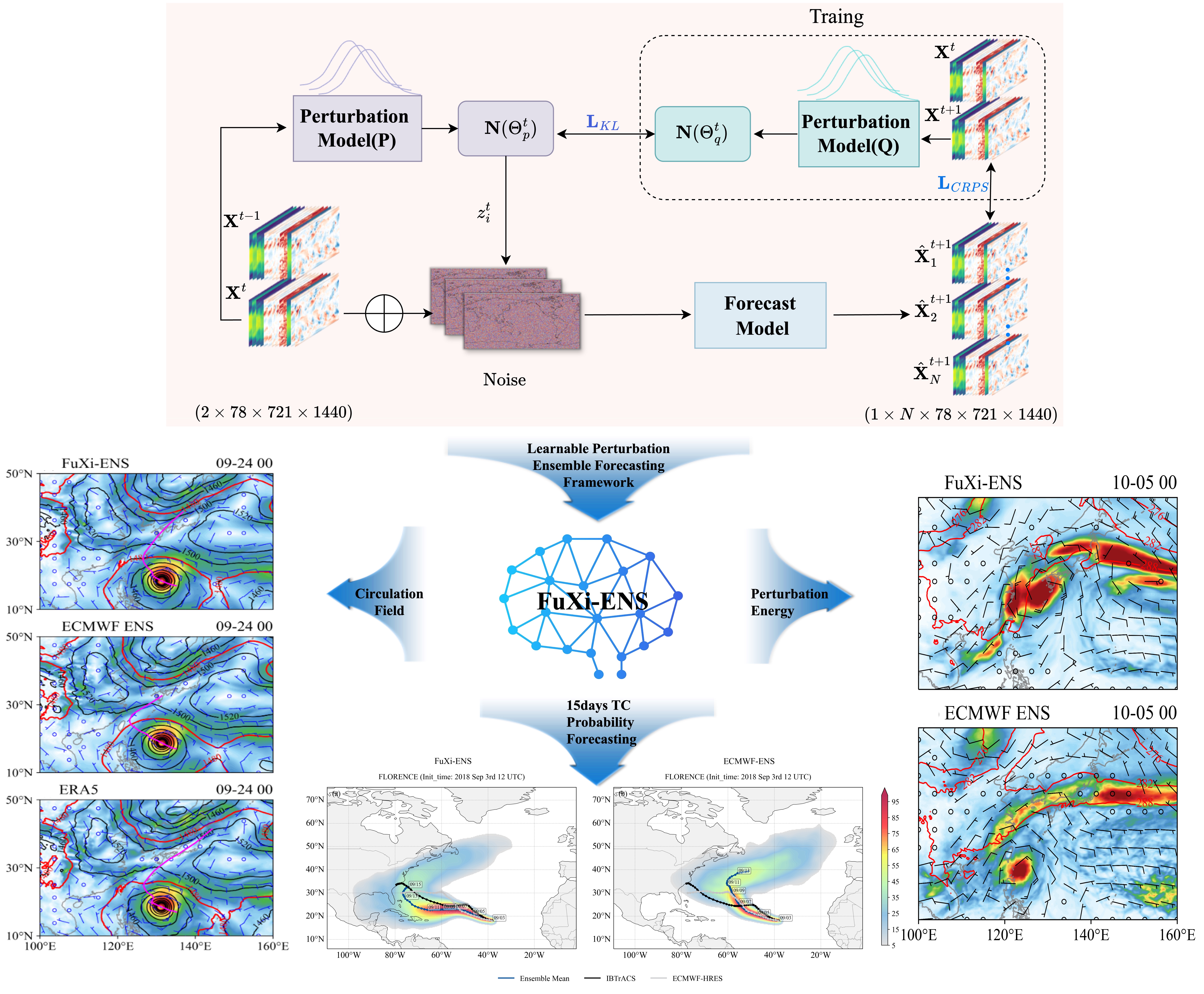}
    \caption{\textbf{Schematic Diagram of Learnable Perturbation Ensemble Forecasting Framework and TC Prediction Performance Evaluation.} The framework includes a perturbation generation module (upper panel) and a forecast model, with evaluation based on atmospheric circulation analysis (left) and perturbation energy assessment (right). Performance is compared between FuXi-ENS, ECMWF-ENS, and ERA5, demonstrating the effectiveness of learnable perturbation methods in TC ensemble prediction.}
    \label{arch}    
\end{figure}
FuXi-ENS adopts a dual-module design architecture. The perturbation generation module is based on a variational autoencoder (VAE) framework \cite{kingma2013auto,kingma2019introduction}. It encodes multi-dimensional meteorological fields into multivariate Gaussian probability distributions. Perturbation vectors are generated through random sampling from these distributions and superimposed onto initial atmospheric states to form perturbed initial fields. The forecasting module employs an encoder-decoder structure that receives perturbed initial fields as input. Combined with  Swin Transformer architecture \cite{Liu2021Swin}, it generates future forecasts through autoregressive processes. The model's most significant technical innovation lies in its flow-dependent perturbation strategy. It introduces perturbations not only at initial time but regenerates flow-dependent perturbations at each forecast time step, superimposing them onto current states. This achieves a dynamic perturbation update mechanism. The design enables the model to adaptively adjust perturbation patterns based on real-time atmospheric evolution characteristics, more accurately capturing uncertainty propagation in nonlinear systems like TCs. FuXi-ENS is trained on ERA5 reanalysis data (2002-2016). Through multiple random sampling and autoregressive forecasting processes, it generates 15-day ensemble forecasts with 48 ensemble members at $0.25^{\circ}$ spatial resolution and 6-hour temporal resolution. The model covers 5 upper-air variables (geopotential (Z), temperature (T), u component of wind (U), v component of wind (V), specific humidity (Q)) and 13 surface variables (2m temperature (T2M), 10m u wind component (U10M), 10m v wind component (V10M), mean sea level pressure (MSL), total precipitation (TP), etc.), providing crucial technical foundation for probabilistic assessment of TC track and intensity prediction. Further details on the FuXi-ENS model can be found in Zhong et al. \cite{zhong2024fuxi}.

\subsection{TC tracking method} \label{fuxi.cascade}


The method employed in this study is adapted from the ECMWF TC tracking algorithm developed by Zhong et al. \cite{zhong2024fuxiextreme,zhang2015verification}. The logic of the tracking algorithm is illustrated in Supplementary Fig. 10. First, a preliminary estimate of the TC position is obtained based on observational data. Within a 445 km radius surrounding this estimated location, candidate points are generated by identifying local minima in MSL and local maxima in 10-meter wind vorticity. Among these candidates, the algorithm selects the nearest point that satisfies a set of predefined criteria (detailed in Supplementary Table 1) as the identified TC position. If none of the candidate points meet the criteria, the TC is considered to be absent and the procedure terminates.
Subsequently, a potential TC position for the next forecast time step is estimated based on the updated position. This estimation combines two approaches: linear extrapolation using the TC positions from the two previous time steps, and the computation of an advection vector derived from wind fields at a specific pressure level. Thereafter, The algorithm applies the same procedure at each time step to identify TC positions.

At the first forecast time step, where only a single historical position is available, the TC position is estimated solely using the advection vector scheme. To mitigate the influence of false local minima caused by the inherent smoothness of machine learning-based forecasts, the ensemble forecast is first downsampled from $0.25^{\circ}$ to $1.25^{\circ}$ before candidate generation and then upsampled back to its original resolution. Furthermore, to avoid excessively large displacements, the TC position change is constrained to no more than three times the displacement in the previous time step.

\subsection{Evaluation metrics}\label{FuXi_ENS}
\textbf{ROCASS:} The Receiver Operating Characteristic (ROC) curve is a widely used metric for evaluating the discriminative ability of binary event forecasts, particularly in probabilistic ensemble prediction \cite{mason2002areas}. By varying the decision threshold, it characterizes the trade-off between the probability of detection (POD) and the probability of false detection (POFD), thereby quantifying the model’s ability to distinguish between \enquote{event} and \enquote{non-event} cases. Plotting POFD on the x-axis and POD on the y-axis yields the ROC curve. The area under the ROC curve (ROCA) provides a numerical summary of forecast skill, ranging from 0 to 1, where values closer to 1 indicate stronger discrimination, and 0.5 denotes no skill beyond random guessing. To facilitate comparison against a no-skill baseline, the ROCA is generally converted into a skill score, referred to as the ROCA skill score (ROCASS) \cite{richardson2000skill}, defined as below:

\begin{equation}
    \mathrm{ROCASS}=2\times(\mathrm{ROCA}-0.5)
\end{equation}

This transformation normalizes the score to the range $[-1, 1]$, where $\text{ROCASS} = 1$ denotes a perfect forecast, $\text{ROCASS} = 0$ indicates equivalence to a no-skill forecast, and $\text{ROCASS} < 0$ implies performance worse than random.

\setlength{\parindent}{0pt}
\setlength{\parskip}{1em}
\textbf{Accumulated Ensemble Mean Position Error and Accumulated Ensemble Spread:} In evaluating TC ensemble track forecasts, it is essential to assess both forecast accuracy and the model's ability to represent forecast uncertainty. Two primary metrics are commonly employed: ensemble mean position error ($\mathrm{Error}{\mathrm{TC}}$)  and ensemble spread ($\mathrm{Spread}{\mathrm{TC}}$). The $\mathrm{Error}{\mathrm{TC}}$, calculated as the root-mean-square error (RMSE) between the ensemble mean track and the observed best track (IBTrACS) \cite{TheRelationshipbetweenEnsembleSpreadandEnsembleMeanSkill}, quantifies the deterministic accuracy of the ensemble mean forecast. The $\mathrm{Spread}{\mathrm{TC}}$, computed as the RMSE between each ensemble member's track and the ensemble mean track, measures the dispersion among members' forecast trajectories. Ideally, $\mathrm{Spread}{\mathrm{TC}}$ should be consistent with the magnitude of $\mathrm{Error}{\mathrm{TC}}$ \cite{yamaguchi2009typhoon}, indicating that the model's uncertainty estimation is reliable. The formulas are as follows:

\setlength{\parskip}{0pt} 


\begin{equation}
    \mathrm{Error_{TC}}(\mathrm{t})=\sqrt{\frac{1}{\mathrm{N}}\sum_{\mathrm{n}=1}^{\mathrm{N}}\left\|\bar{\mathrm{p}}_\mathrm{forecast}^{(\mathrm{n,t})}-\mathrm{p}_{\mathrm{o}}^{(\mathrm{n,t})}\right\|^2}
\end{equation}

\begin{equation}
    \mathrm{Spread_{TC}}(\mathrm{t})=\sqrt{\frac{1}{\mathrm{N}}\sum_{\mathrm{n}=1}^{\mathrm{N}}\frac{1}{\mathrm{M}}\sum_{\mathrm{m}=1}^{\mathrm{M}}\left\|\mathrm{p}_{\mathrm{forecast}}^{(\mathrm{n,m,t})}-\bar{\mathrm{p}}_{\mathrm{forecast}}^{(\mathrm{n,t})}\right\|^2}
\end{equation}


where $\mathrm{N}$ represents the number of samples, $\mathrm{M}$ denotes the number of ensemble members, $\bar{\mathrm{p}}_{\text{forecast}}^{(\mathrm{n,t})}$ represents the ensemble mean position for the $\mathrm{n}^{\mathrm{th}}$ sample at forecast lead time $\mathrm{t}$, $\mathrm{p}_{\mathrm{o}}^{(\mathrm{n,t})}$ represents the observed best track position for the $\mathrm{n}^{\mathrm{th}}$ sample at lead time $\mathrm{t}$, and $\mathrm{p}_{\text{forecast}}^{(\mathrm{n,m,t})}$ represents the forecast position for the $\mathrm{n}^{\mathrm{th}}$ sample, $\mathrm{m}^{\mathrm{th}}$ member at lead time $\mathrm{t}$.

\setlength{\parindent}{15pt}  
By accumulating $\mathrm{Error}{\mathrm{TC}}$ and $\mathrm{Spread}{\mathrm{TC}}$ across multiple forecast lead times, the Accumulated Ensemble Mean Position Error ($\mathrm{AccError}_{\mathrm{TC}}$) and Accumulated Ensemble Spread ($\mathrm{AccSpread}_{\mathrm{TC}}$) are obtained. These accumulated metrics are used to comprehensively evaluate the track prediction accuracy and overall uncertainty throughout the forecast period:


\begin{equation}
    \mathrm{AccError}_{\mathrm{TC}}=\sum_{\mathrm{t}\in \mathrm{T}}\mathrm{Error}_{\mathrm{TC}}(\mathrm{t})
\end{equation}

\begin{equation}
    \mathrm{AccSpread}_{\mathrm{TC}}=\sum_{\mathrm{t}\in \mathrm{T}}\mathrm{Spread}_{\mathrm{TC}}(\mathrm{t})
\end{equation}

Ideally, $\mathrm{AccSpread}_{\mathrm{TC}}$ should match $\mathrm{AccError}_{\mathrm{TC}}$. If $\mathrm{AccSpread}_{\mathrm{TC}} > \mathrm{AccERROR}_{\mathrm{TC}}$, it indicates an overestimation of forecast uncertainty. If $\mathrm{AccSpread}_{\mathrm{TC}} < \mathrm{AccERROR}_{\mathrm{TC}}$, it signifies an underestimation of forecast uncertainty.

\setlength{\parindent}{0pt}
\setlength{\parskip}{1em}
\textbf{Along-track Error and Cross-track Error:} To quantify TC track forcast errors, this study decomposes $\mathrm{Error}{\mathrm{TC}}$ into Along-track error (AT) and Cross-track error (CT) \cite{goerss2000tropical,aberson2001ensemble}. The calculation proceeds as follows (The detailed schematic diagram of the calculation principle is shown in Supplementary Fig. 11).
Let two consecutive observed positions be denoted as $\mathrm{OB}_1$ and $\mathrm{OB}_2$, and the forecast position as FC. AT represents the displacement of the forecast point in the tangential direction of the observed track and is defined as \cite{cangialosi2012national,goerss2004history}:


\setlength{\parskip}{0pt}

\begin{equation}
\mathrm{AT} = \frac{\overrightarrow{\mathrm{OB_1OB_2}} \cdot \overrightarrow{\mathrm{OB_1FC}}}{\lvert\lvert\overrightarrow{\mathrm{OB_1OB_2}}\rvert\rvert} - \lvert\lvert\overrightarrow{\mathrm{OB_1OB_2}}\rvert\rvert
\end{equation}

\setlength{\parindent}{0pt}
where $\overrightarrow{\mathrm{OB_1OB_2}}$ is the track tangent vector, and $\overrightarrow{\mathrm{OB_1FC}}$ is the vector from $\mathrm{OB_1}$ to the forecast point. $\mathrm{AT}$ is defined as the projection error of the forecast position in the direction of the observed track: $\mathrm{AT} > 0$ indicates the forecast position is ahead of the observed track, $\mathrm{AT} < 0$ indicates the forecast position lags behind the observed track.

\setlength{\parindent}{15pt}
CT measures the displacement perpendicular to the observed track and is computed using the Pythagorean relation:


\begin{equation}
    \mathrm{CT}=\sqrt{\lvert\lvert\overrightarrow{\mathrm{OB_{1}FC}}\rvert\rvert^2-\left(\frac{\overrightarrow{\mathrm{OB_{1}OB_{2}}}\cdot\overrightarrow{\mathrm{OB_{1}FC}}}{\lvert\lvert\overrightarrow{\mathrm{OB_{1}OB_{2}}}\rvert\rvert}\right)^2}
\end{equation}

 $\mathrm{CT}$ is defined as the deviation of the forecast position perpendicular to the observed track direction: $\mathrm{CT} > 0$ indicates that the forecast position lies to the right of the observed track, while $\mathrm{CT} < 0$ indicates that it lies to the left. Together, $\mathrm{AT}$ and $\mathrm{CT}$ provide a clear decomposition of forecast error into tangential and normal components, forming a quantitative basis for evaluating $\mathrm{TC}$ track prediction performance.

\setlength{\parindent}{0pt}
\setlength{\parskip}{1em}

\textbf{Mann-Whitney U test:} To evaluate the statistical significance of performance differences between the two ensemble forecasting systems in TC prediction, we employed the Mann-Whitney U test. This non-parametric test is particularly well-suited for comparing TC forecast skill metrics between the ECMWF-ENS and FuXi-ENS systems, as it does not require the assumption of normality—a condition frequently violated by meteorological forecast verification scores. The Mann-Whitney U test assesses whether two independent samples originate from the same underlying distribution by comparing their rank-ordered values, thereby providing robustness against outliers that may occur in TC forecast performance data \cite{kirkland2019regional,nachar2008mann}.

\textbf{Strike Probability:} To quantify the uncertainty in TC forecasts, we calculate the strike probability using ensemble predictions. The computational method can be expressed as follows \cite{strikeprobability, strikeprobability2}:

\setlength{\parskip}{0pt}
\setlength{\parindent}{15pt}
The strike probability for grid point $(\mathrm{i},\mathrm{j})$ is calculated as:
\begin{equation}
\mathrm{P}(\mathrm{i},\mathrm{j}) = \frac{1}{\mathrm{N}} \times \sum_{\mathrm{k}=1}^{\mathrm{N}} \mathrm{I}_{\mathrm{k}}(\mathrm{i},\mathrm{j}) \times 100\%
\end{equation}

\setlength{\parindent}{0pt}
where $\mathrm{N}$ is the total ensemble members. the impact zone indicator function $\mathrm{I}_{\mathrm{k}}$ is defined as:
\begin{equation}
\mathrm{I}_{\mathrm{k}}(\mathrm{i},\mathrm{j}) = \begin{cases} 
1, & \text{if } \mathrm{d}(\mathrm{point}(\mathrm{i},\mathrm{j}), \mathrm{track}_{\mathrm{k}}) \leq \mathrm{R} \\
0, & \text{otherwise}
\end{cases}
\end{equation}

where $\mathrm{d}$ is the minimum distance from point to track; $\mathrm{R}$ is the impact radius (1°, approximately 111 km).

\setlength{\parindent}{15pt}
For multiple concurrent TCs, a maximum value merging strategy is applied:
\begin{equation}
\mathrm{P}_{\mathrm{total}}(\mathrm{i},\mathrm{j}) = \max\{\mathrm{P}_{1}(\mathrm{i},\mathrm{j}), \mathrm{P}_{2}(\mathrm{i},\mathrm{j}), \ldots, \mathrm{P}_{\mathrm{m}}(\mathrm{i},\mathrm{j})\}
\end{equation}

\setlength{\parindent}{0pt}
where $\mathrm{m}$ is the number of concurrent TCs.

\setlength{\parindent}{0pt}
\setlength{\parskip}{1em}
\textbf{Moisture Turbulent Energy：}To quantitatively assess the energy characteristics of perturbations in ensemble forecasts, we adopt a thermodynamic perspective of TC processes and employ Moisture Turbulent Energy (MTE) perturbations as a diagnostic metric. MTE comprehensively incorporates perturbation components of wind speed, temperature, and specific humidity, effectively reflecting the perturbation characteristics of atmospheric thermodynamic states within TC systems. The calculation formula is \cite{shao2024impact}:


\begin{equation}
    \begin{aligned}\mathrm{MTE_{TC}}=&\frac{1}{2}\left[{u^{\prime}}^2_{\mathrm{i,j,k}}+{v^{\prime}}^2_{\mathrm{i,j,k}}\right]+&\frac{\mathrm{c_p}}{\mathrm{T_r}}{\mathrm{T^{\prime}}}^2_{\mathrm{i,j,k}}+\varepsilon\frac{\mathrm{L}^2}{\mathrm{c_p}\mathrm{T_r}}{\mathrm{q^{\prime}}}^2_{\mathrm{i,j,k}}\end{aligned}
\end{equation}

where $\mathrm{u^{\prime}}$ and $\mathrm{v^{\prime}}$ denote the zonal and meridional wind perturbation respectively, $\mathrm{T^{\prime}}$ represents the temperature perturbation, and $\mathrm{q^{\prime}}$ is the specific humidity perturbation. Each perturbation is defined as the deviation of an ensemble member forecast from the ensemble mean. The constants are specified as follows: reference temperature $\mathrm{T_{r}}=270\;\mathrm{K}$, specific heat of dry air at constant pressure $\mathrm{c_{p}}=1005.7\;\mathrm{J\,kg^{-1}\,K^{-1}}$, and latent heat of water vapor condensation $\mathrm{L}=2.51\times10^{6}\;\mathrm{J\,kg^{-1}}$.

\setlength{\parindent}{15pt}
\setlength{\parskip}{0pt}
In this formulation, the first term represents the contribution from perturbation kinetic energy, the second term reflects the effect of temperature perturbations on available potential energy, and the third term captures the enhancement of perturbation energy through latent heat release associated with specific humidity. Compared with traditional dry-air perturbation energy measures, $\mathrm{MTE}$ provides a more comprehensive description of the role of moist processes in the growth of ensemble perturbations. In practice, $\mathrm{MTE}$ is evaluated at each horizontal grid point $(\mathrm{i}, \mathrm{j})$ and vertical level $\mathrm{k}$, thereby yielding a three-dimensional distribution of perturbation energy.

\pagestyle{plain} 

\pagestyle{plain} 
\section*{Data availability}

We downloaded a subset of the ERA5 dataset from the official Copernicus Climate Data Store (CDS) at \url{https://cds.climate.copernicus.eu/}. Additionally, ground truth TC tracks were obtained from the International Best Track Archive for Climate Stewardship (IBTrACS) project, publicly available at \url{https://www.ncei.noaa.gov/products/international-best-track-archive}. TC track forecasts from ECMWF-HRES (deterministic) and ECMWF-ENS (ensemble) were retrieved from the TIGGE (THORPEX Interactive Grand Global Ensemble) data archive at \url{https://confluence.ecmwf.int/display/TIGGE/Tools}. Furthermore, the historical operational ensemble forecast data used in this study were accessed from the WeatherBench2 platform's Google Cloud storage service at \url{gs://weatherbench2/datasets/ifs_ens/2018-2022-1440x721.zarr}.

\pagestyle{plain} 
\section*{Code availability} 

The FuXi-ENS model and related example code used for evaluation in this study are publicly available and can be accessed from Zenodo:\url{https://zenodo.org/records/15124541/files/FuXi-ENS-main.zip?download=1}.


\pagestyle{plain}
\section*{Acknowledgements}

We are grateful to ECMWF for providing the ERA5 dataset and acknowledge the ensemble forecast data from WeatherBench2 available through Google Cloud. We express our gratitude to the researchers at ECMWF for providing the ERA5 dataset and HRES to the research community. We acknowledged the efforts of NOAA National Centers for Environmental Information in making the IBTrACS dataset available. The computations in this research were performed using the CFFF platform of Fudan University. We also thank support from the Computing for the Future at Fudan (CFFF). We are grateful to Shanghai Academy of Artificial Intelligence for Science (SAIS) for providing the internship opportunity, and this work was primarily completed during the internship period. 

\pagestyle{plain}
\section*{Author Contributions}
Jun Liu designed the experiments, processed the data, performed evaluation analyses and calculations, and wrote the manuscript. Tao Zhou conducted the circulation field analysis. Jiarui Li wrote the evaluation  metrics sections of the manuscript. Xiaohui Zhong, Peng Zhang and Jie Feng contributed to manuscript review. Xiaohui Zhong, Lei Chen, and Hao Li developed the FuXi-ENS model. Hao Li supervised the project and provided overall planning, management, and guidance.
\pagestyle{plain} 
\section*{Ethics declarations}
The authors declare no competing interests.

\pagestyle{plain}  
\bibliography{ensemble_tropical_cyclones_refs}


\begin{thebibliography}{72}
\ifx \bisbn   \undefined \def \bisbn  #1{ISBN #1}\fi
\ifx \binits  \undefined \def \binits#1{#1}\fi
\ifx \bauthor  \undefined \def \bauthor#1{#1}\fi
\ifx \batitle  \undefined \def \batitle#1{#1}\fi
\ifx \bjtitle  \undefined \def \bjtitle#1{#1}\fi
\ifx \bvolume  \undefined \def \bvolume#1{\textbf{#1}}\fi
\ifx \byear  \undefined \def \byear#1{#1}\fi
\ifx \bissue  \undefined \def \bissue#1{#1}\fi
\ifx \bfpage  \undefined \def \bfpage#1{#1}\fi
\ifx \blpage  \undefined \def \blpage #1{#1}\fi
\ifx \burl  \undefined \def \burl#1{\textsf{#1}}\fi
\ifx \doiurl  \undefined \def \doiurl#1{\url{https://doi.org/#1}}\fi
\ifx \betal  \undefined \def \betal{\textit{et al.}}\fi
\ifx \binstitute  \undefined \def \binstitute#1{#1}\fi
\ifx \binstitutionaled  \undefined \def \binstitutionaled#1{#1}\fi
\ifx \bctitle  \undefined \def \bctitle#1{#1}\fi
\ifx \beditor  \undefined \def \beditor#1{#1}\fi
\ifx \bpublisher  \undefined \def \bpublisher#1{#1}\fi
\ifx \bbtitle  \undefined \def \bbtitle#1{#1}\fi
\ifx \bedition  \undefined \def \bedition#1{#1}\fi
\ifx \bseriesno  \undefined \def \bseriesno#1{#1}\fi
\ifx \blocation  \undefined \def \blocation#1{#1}\fi
\ifx \bsertitle  \undefined \def \bsertitle#1{#1}\fi
\ifx \bsnm \undefined \def \bsnm#1{#1}\fi
\ifx \bsuffix \undefined \def \bsuffix#1{#1}\fi
\ifx \bparticle \undefined \def \bparticle#1{#1}\fi
\ifx \barticle \undefined \def \barticle#1{#1}\fi
\bibcommenthead
\ifx \bconfdate \undefined \def \bconfdate #1{#1}\fi
\ifx \botherref \undefined \def \botherref #1{#1}\fi
\ifx \url \undefined \def \url#1{\textsf{#1}}\fi
\ifx \bchapter \undefined \def \bchapter#1{#1}\fi
\ifx \bbook \undefined \def \bbook#1{#1}\fi
\ifx \bcomment \undefined \def \bcomment#1{#1}\fi
\ifx \oauthor \undefined \def \oauthor#1{#1}\fi
\ifx \citeauthoryear \undefined \def \citeauthoryear#1{#1}\fi
\ifx \endbibitem  \undefined \def \endbibitem {}\fi
\ifx \bconflocation  \undefined \def \bconflocation#1{#1}\fi
\ifx \arxivurl  \undefined \def \arxivurl#1{\textsf{#1}}\fi
\csname PreBibitemsHook\endcsname

\bibitem{emanuel2005increasing}
\begin{barticle}
\bauthor{\bsnm{Emanuel}, \binits{K.}}:
\batitle{Increasing destructiveness of tropical cyclones over the past 30 years}.
\bjtitle{Nature}
\bvolume{436}(\bissue{7051}),
\bfpage{686}--\blpage{688}
(\byear{2005})
\end{barticle}
\endbibitem

\bibitem{mendelsohn2012impact}
\begin{barticle}
\bauthor{\bsnm{Mendelsohn}, \binits{R.}},
\bauthor{\bsnm{Emanuel}, \binits{K.}},
\bauthor{\bsnm{Chonabayashi}, \binits{S.}},
\bauthor{\bsnm{Bakkensen}, \binits{L.}}:
\batitle{The impact of climate change on global tropical cyclone damage}.
\bjtitle{Nature climate change}
\bvolume{2}(\bissue{3}),
\bfpage{205}--\blpage{209}
(\byear{2012})
\end{barticle}
\endbibitem

\bibitem{peduzzi2012global}
\begin{barticle}
\bauthor{\bsnm{Peduzzi}, \binits{P.}},
\bauthor{\bsnm{Chatenoux}, \binits{B.}},
\bauthor{\bsnm{Dao}, \binits{H.}},
\bauthor{\bsnm{De~Bono}, \binits{A.}},
\bauthor{\bsnm{Herold}, \binits{C.}},
\bauthor{\bsnm{Kossin}, \binits{J.}},
\bauthor{\bsnm{Mouton}, \binits{F.}},
\bauthor{\bsnm{Nordbeck}, \binits{O.}}:
\batitle{Global trends in tropical cyclone risk}.
\bjtitle{Nature climate change}
\bvolume{2}(\bissue{4}),
\bfpage{289}--\blpage{294}
(\byear{2012})
\end{barticle}
\endbibitem

\bibitem{zhang2009tropical}
\begin{barticle}
\bauthor{\bsnm{Zhang}, \binits{Q.}},
\bauthor{\bsnm{Wu}, \binits{L.}},
\bauthor{\bsnm{Liu}, \binits{Q.}}:
\batitle{Tropical cyclone damages in china 1983--2006}.
\bjtitle{Bulletin of the American Meteorological Society}
\bvolume{90}(\bissue{4}),
\bfpage{489}--\blpage{496}
(\byear{2009})
\end{barticle}
\endbibitem

\bibitem{doocy2013human}
\begin{barticle}
\bauthor{\bsnm{Doocy}, \binits{S.}},
\bauthor{\bsnm{Dick}, \binits{A.}},
\bauthor{\bsnm{Daniels}, \binits{A.}},
\bauthor{\bsnm{Kirsch}, \binits{T.D.}}:
\batitle{The human impact of tropical cyclones: a historical review of events 1980-2009 and systematic literature review}.
\bjtitle{PLoS currents}
\bvolume{5},
(\byear{2013})
\end{barticle}
\endbibitem

\bibitem{qin2024recent}
\begin{barticle}
\bauthor{\bsnm{Qin}, \binits{L.}},
\bauthor{\bsnm{Zhu}, \binits{L.}},
\bauthor{\bsnm{Liao}, \binits{X.}},
\bauthor{\bsnm{Meng}, \binits{C.}},
\bauthor{\bsnm{Han}, \binits{Q.}},
\bauthor{\bsnm{Li}, \binits{Z.}},
\bauthor{\bsnm{Shen}, \binits{S.}},
\bauthor{\bsnm{Xu}, \binits{W.}},
\bauthor{\bsnm{Chen}, \binits{J.}}:
\batitle{Recent northward shift of tropical cyclone economic risk in china}.
\bjtitle{npj Natural Hazards}
\bvolume{1}(\bissue{1}),
\bfpage{8}
(\byear{2024})
\end{barticle}
\endbibitem

\bibitem{palmer2000predicting}
\begin{barticle}
\bauthor{\bsnm{Palmer}, \binits{T.N.}}:
\batitle{Predicting uncertainty in forecasts of weather and climate}.
\bjtitle{Reports on progress in Physics}
\bvolume{63}(\bissue{2}),
\bfpage{71}
(\byear{2000})
\end{barticle}
\endbibitem

\bibitem{lang2024aifs}
\begin{botherref}
\oauthor{\bsnm{Lang}, \binits{S.}},
\oauthor{\bsnm{Alexe}, \binits{M.}},
\oauthor{\bsnm{Clare}, \binits{M.C.}},
\oauthor{\bsnm{Roberts}, \binits{C.}},
\oauthor{\bsnm{Adewoyin}, \binits{R.}},
\oauthor{\bsnm{Bouall{\`e}gue}, \binits{Z.B.}},
\oauthor{\bsnm{Chantry}, \binits{M.}},
\oauthor{\bsnm{Dramsch}, \binits{J.}},
\oauthor{\bsnm{Dueben}, \binits{P.D.}},
\oauthor{\bsnm{Hahner}, \binits{S.}}, et al.:
Aifs-crps: ensemble forecasting using a model trained with a loss function based on the continuous ranked probability score.
arXiv preprint arXiv:2412.15832
(2024)
\end{botherref}
\endbibitem

\bibitem{leutbecher2008ensemble}
\begin{barticle}
\bauthor{\bsnm{Leutbecher}, \binits{M.}},
\bauthor{\bsnm{Palmer}, \binits{T.N.}}:
\batitle{Ensemble forecasting}.
\bjtitle{Journal of computational physics}
\bvolume{227}(\bissue{7}),
\bfpage{3515}--\blpage{3539}
(\byear{2008})
\end{barticle}
\endbibitem

\bibitem{feng2024ensemble}
\begin{barticle}
\bauthor{\bsnm{Feng}, \binits{J.}},
\bauthor{\bsnm{Toth}, \binits{Z.}},
\bauthor{\bsnm{Zhang}, \binits{J.}},
\bauthor{\bsnm{Pe{\~n}a}, \binits{M.}}:
\batitle{Ensemble forecasting: a foray of dynamics into the realm of statistics}.
\bjtitle{Quarterly Journal of the Royal Meteorological Society}
\bvolume{150}(\bissue{762}),
\bfpage{2537}--\blpage{2560}
(\byear{2024})
\end{barticle}
\endbibitem

\bibitem{wang2020uncertainty}
\begin{barticle}
\bauthor{\bsnm{Wang}, \binits{C.}},
\bauthor{\bsnm{Zeng}, \binits{Z.}},
\bauthor{\bsnm{Ying}, \binits{M.}}:
\batitle{Uncertainty in tropical cyclone intensity predictions due to uncertainty in initial conditions}.
\bjtitle{Advances in Atmospheric Sciences}
\bvolume{37}(\bissue{3}),
\bfpage{278}--\blpage{290}
(\byear{2020})
\end{barticle}
\endbibitem

\bibitem{zhang2017impact}
\begin{barticle}
\bauthor{\bsnm{Zhang}, \binits{Q.}},
\bauthor{\bsnm{Gu}, \binits{X.}},
\bauthor{\bsnm{Shi}, \binits{P.}},
\bauthor{\bsnm{Singh}, \binits{V.P.}}:
\batitle{Impact of tropical cyclones on flood risk in southeastern china: Spatial patterns, causes and implications}.
\bjtitle{Global and Planetary Change}
\bvolume{150},
\bfpage{81}--\blpage{93}
(\byear{2017})
\end{barticle}
\endbibitem

\bibitem{buizza1999stochastic}
\begin{barticle}
\bauthor{\bsnm{Buizza}, \binits{R.}},
\bauthor{\bsnm{Milleer}, \binits{M.}},
\bauthor{\bsnm{Palmer}, \binits{T.N.}}:
\batitle{Stochastic representation of model uncertainties in the ecmwf ensemble prediction system}.
\bjtitle{Quarterly Journal of the Royal Meteorological Society}
\bvolume{125}(\bissue{560}),
\bfpage{2887}--\blpage{2908}
(\byear{1999})
\end{barticle}
\endbibitem

\bibitem{buizza2019introduction}
\begin{barticle}
\bauthor{\bsnm{Buizza}, \binits{R.}}:
\batitle{Introduction to the special issue on “25 years of ensemble forecasting”}.
\bjtitle{Quarterly Journal of the Royal Meteorological Society}
\bvolume{145},
\bfpage{1}--\blpage{11}
(\byear{2019})
\end{barticle}
\endbibitem

\bibitem{buizza1994localization}
\begin{barticle}
\bauthor{\bsnm{Buizza}, \binits{R.}}:
\batitle{Localization of optimal perturbations using a projection operator}.
\bjtitle{Quarterly Journal of the Royal Meteorological Society}
\bvolume{120}(\bissue{520}),
\bfpage{1647}--\blpage{1681}
(\byear{1994})
\end{barticle}
\endbibitem

\bibitem{buizza1995singular}
\begin{barticle}
\bauthor{\bsnm{Buizza}, \binits{R.}},
\bauthor{\bsnm{Palmer}, \binits{T.N.}}:
\batitle{The singular-vector structure of the atmospheric global circulation}.
\bjtitle{Journal of Atmospheric Sciences}
\bvolume{52}(\bissue{9}),
\bfpage{1434}--\blpage{1456}
(\byear{1995})
\end{barticle}
\endbibitem

\bibitem{buizza2008potential}
\begin{barticle}
\bauthor{\bsnm{Buizza}, \binits{R.}},
\bauthor{\bsnm{Leutbecher}, \binits{M.}},
\bauthor{\bsnm{Isaksen}, \binits{L.}}:
\batitle{Potential use of an ensemble of analyses in the ecmwf ensemble prediction system}.
\bjtitle{Quarterly Journal of the Royal Meteorological Society: A journal of the atmospheric sciences, applied meteorology and physical oceanography}
\bvolume{134}(\bissue{637}),
\bfpage{2051}--\blpage{2066}
(\byear{2008})
\end{barticle}
\endbibitem

\bibitem{wastl2019independent}
\begin{barticle}
\bauthor{\bsnm{Wastl}, \binits{C.}},
\bauthor{\bsnm{Wang}, \binits{Y.}},
\bauthor{\bsnm{Atencia}, \binits{A.}},
\bauthor{\bsnm{Wittmann}, \binits{C.}}:
\batitle{Independent perturbations for physics parametrization tendencies in a convection-permitting ensemble (psppt)}.
\bjtitle{Geoscientific Model Development}
\bvolume{12}(\bissue{1}),
\bfpage{261}--\blpage{273}
(\byear{2019})
\end{barticle}
\endbibitem

\bibitem{langland2012recent}
\begin{barticle}
\bauthor{\bsnm{Langland}, \binits{R.H.}},
\bauthor{\bsnm{Maue}, \binits{R.N.}}:
\batitle{Recent northern hemisphere mid-latitude medium-range deterministic forecast skill}.
\bjtitle{Tellus A: Dynamic Meteorology and Oceanography}
\bvolume{64}(\bissue{1}),
\bfpage{17531}
(\byear{2012})
\end{barticle}
\endbibitem

\bibitem{molteni1996ecmwf}
\begin{barticle}
\bauthor{\bsnm{Molteni}, \binits{F.}},
\bauthor{\bsnm{Buizza}, \binits{R.}},
\bauthor{\bsnm{Palmer}, \binits{T.N.}},
\bauthor{\bsnm{Petroliagis}, \binits{T.}}:
\batitle{The ecmwf ensemble prediction system: Methodology and validation}.
\bjtitle{Quarterly journal of the royal meteorological society}
\bvolume{122}(\bissue{529}),
\bfpage{73}--\blpage{119}
(\byear{1996})
\end{barticle}
\endbibitem

\bibitem{palmer2019ecmwf}
\begin{barticle}
\bauthor{\bsnm{Palmer}, \binits{T.}}:
\batitle{The ecmwf ensemble prediction system: Looking back (more than) 25 years and projecting forward 25 years}.
\bjtitle{Quarterly Journal of the Royal Meteorological Society}
\bvolume{145},
\bfpage{12}--\blpage{24}
(\byear{2019})
\end{barticle}
\endbibitem

\bibitem{aijaz2019bias}
\begin{barticle}
\bauthor{\bsnm{Aijaz}, \binits{S.}},
\bauthor{\bsnm{Kepert}, \binits{J.D.}},
\bauthor{\bsnm{Ye}, \binits{H.}},
\bauthor{\bsnm{Huang}, \binits{Z.}},
\bauthor{\bsnm{Hawksford}, \binits{A.}}:
\batitle{Bias correction of tropical cyclone parameters in the ecmwf ensemble prediction system in australia}.
\bjtitle{Monthly Weather Review}
\bvolume{147}(\bissue{11}),
\bfpage{4261}--\blpage{4285}
(\byear{2019})
\end{barticle}
\endbibitem

\bibitem{elsberry2021predicting}
\begin{barticle}
\bauthor{\bsnm{Elsberry}, \binits{R.L.}},
\bauthor{\bsnm{Tsai}, \binits{H.-C.}},
\bauthor{\bsnm{Chin}, \binits{W.-C.}},
\bauthor{\bsnm{Marchok}, \binits{T.P.}}:
\batitle{Predicting rapid intensification events following tropical cyclone formation in the western north pacific based on ecmwf ensemble warm core evolutions}.
\bjtitle{Atmosphere}
\bvolume{12}(\bissue{7}),
\bfpage{847}
(\byear{2021})
\end{barticle}
\endbibitem

\bibitem{hamill2013noaa}
\begin{barticle}
\bauthor{\bsnm{Hamill}, \binits{T.M.}},
\bauthor{\bsnm{Bates}, \binits{G.T.}},
\bauthor{\bsnm{Whitaker}, \binits{J.S.}},
\bauthor{\bsnm{Murray}, \binits{D.R.}},
\bauthor{\bsnm{Fiorino}, \binits{M.}},
\bauthor{\bsnm{Galarneau~Jr}, \binits{T.J.}},
\bauthor{\bsnm{Zhu}, \binits{Y.}},
\bauthor{\bsnm{Lapenta}, \binits{W.}}:
\batitle{Noaa's second-generation global medium-range ensemble reforecast dataset}.
\bjtitle{Bulletin of the American Meteorological Society}
\bvolume{94}(\bissue{10}),
\bfpage{1553}--\blpage{1565}
(\byear{2013})
\end{barticle}
\endbibitem

\bibitem{houtekamer2001sequential}
\begin{barticle}
\bauthor{\bsnm{Houtekamer}, \binits{P.L.}},
\bauthor{\bsnm{Mitchell}, \binits{H.L.}}:
\batitle{A sequential ensemble kalman filter for atmospheric data assimilation}.
\bjtitle{Monthly weather review}
\bvolume{129}(\bissue{1}),
\bfpage{123}--\blpage{137}
(\byear{2001})
\end{barticle}
\endbibitem

\bibitem{bishop2001adaptive}
\begin{barticle}
\bauthor{\bsnm{Bishop}, \binits{C.H.}},
\bauthor{\bsnm{Etherton}, \binits{B.J.}},
\bauthor{\bsnm{Majumdar}, \binits{S.J.}}:
\batitle{Adaptive sampling with the ensemble transform kalman filter. part i: Theoretical aspects}.
\bjtitle{Monthly weather review}
\bvolume{129}(\bissue{3}),
\bfpage{420}--\blpage{436}
(\byear{2001})
\end{barticle}
\endbibitem

\bibitem{zhang2009cloud}
\begin{barticle}
\bauthor{\bsnm{Zhang}, \binits{F.}},
\bauthor{\bsnm{Weng}, \binits{Y.}},
\bauthor{\bsnm{Sippel}, \binits{J.A.}},
\bauthor{\bsnm{Meng}, \binits{Z.}},
\bauthor{\bsnm{Bishop}, \binits{C.H.}}:
\batitle{Cloud-resolving hurricane initialization and prediction through assimilation of doppler radar observations with an ensemble kalman filter}.
\bjtitle{Monthly Weather Review}
\bvolume{137}(\bissue{7}),
\bfpage{2105}--\blpage{2125}
(\byear{2009})
\end{barticle}
\endbibitem

\bibitem{zhou2019toward}
\begin{barticle}
\bauthor{\bsnm{Zhou}, \binits{L.}},
\bauthor{\bsnm{Lin}, \binits{S.-J.}},
\bauthor{\bsnm{Chen}, \binits{J.-H.}},
\bauthor{\bsnm{Harris}, \binits{L.M.}},
\bauthor{\bsnm{Chen}, \binits{X.}},
\bauthor{\bsnm{Rees}, \binits{S.L.}}:
\batitle{Toward convective-scale prediction within the next generation global prediction system}.
\bjtitle{Bulletin of the American Meteorological Society}
\bvolume{100}(\bissue{7}),
\bfpage{1225}--\blpage{1243}
(\byear{2019})
\end{barticle}
\endbibitem

\bibitem{hazelton2018evaluation}
\begin{barticle}
\bauthor{\bsnm{Hazelton}, \binits{A.T.}},
\bauthor{\bsnm{Harris}, \binits{L.}},
\bauthor{\bsnm{Lin}, \binits{S.-J.}}:
\batitle{Evaluation of tropical cyclone structure forecasts in a high-resolution version of the multiscale gfdl fvgfs model}.
\bjtitle{Weather and Forecasting}
\bvolume{33}(\bissue{2}),
\bfpage{419}--\blpage{442}
(\byear{2018})
\end{barticle}
\endbibitem

\bibitem{zhang2009coupling}
\begin{barticle}
\bauthor{\bsnm{Zhang}, \binits{F.}},
\bauthor{\bsnm{Zhang}, \binits{M.}},
\bauthor{\bsnm{Hansen}, \binits{J.A.}}:
\batitle{Coupling ensemble kalman filter with four-dimensional variational data assimilation}.
\bjtitle{Advances in Atmospheric Sciences}
\bvolume{26}(\bissue{1}),
\bfpage{1}--\blpage{8}
(\byear{2009})
\end{barticle}
\endbibitem

\bibitem{zhang2012e4dvar}
\begin{barticle}
\bauthor{\bsnm{Zhang}, \binits{M.}},
\bauthor{\bsnm{Zhang}, \binits{F.}}:
\batitle{E4dvar: Coupling an ensemble kalman filter with four-dimensional variational data assimilation in a limited-area weather prediction model}.
\bjtitle{Monthly Weather Review}
\bvolume{140}(\bissue{2}),
\bfpage{587}--\blpage{600}
(\byear{2012})
\end{barticle}
\endbibitem

\bibitem{buizza2005comparison}
\begin{barticle}
\bauthor{\bsnm{Buizza}, \binits{R.}},
\bauthor{\bsnm{Houtekamer}, \binits{P.}},
\bauthor{\bsnm{Pellerin}, \binits{G.}},
\bauthor{\bsnm{Toth}, \binits{Z.}},
\bauthor{\bsnm{Zhu}, \binits{Y.}},
\bauthor{\bsnm{Wei}, \binits{M.}}:
\batitle{A comparison of the ecmwf, msc, and ncep global ensemble prediction systems}.
\bjtitle{Monthly Weather Review}
\bvolume{133}(\bissue{5}),
\bfpage{1076}--\blpage{1097}
(\byear{2005})
\end{barticle}
\endbibitem

\bibitem{bauer2015quiet}
\begin{barticle}
\bauthor{\bsnm{Bauer}, \binits{P.}},
\bauthor{\bsnm{Thorpe}, \binits{A.}},
\bauthor{\bsnm{Brunet}, \binits{G.}}:
\batitle{The quiet revolution of numerical weather prediction}.
\bjtitle{Nature}
\bvolume{525}(\bissue{7567}),
\bfpage{47}--\blpage{55}
(\byear{2015})
\end{barticle}
\endbibitem

\bibitem{pu2025fast}
\begin{barticle}
\bauthor{\bsnm{Pu}, \binits{J.}},
\bauthor{\bsnm{Mu}, \binits{M.}},
\bauthor{\bsnm{Feng}, \binits{J.}},
\bauthor{\bsnm{Zhong}, \binits{X.}},
\bauthor{\bsnm{Li}, \binits{H.}}:
\batitle{A fast physics-based perturbation generator of machine learning weather model for efficient ensemble forecasts of tropical cyclone track}.
\bjtitle{npj Climate and Atmospheric Science}
\bvolume{8}(\bissue{1}),
\bfpage{128}
(\byear{2025})
\end{barticle}
\endbibitem

\bibitem{pathak2022fourcastnet}
\begin{botherref}
\oauthor{\bsnm{Pathak}, \binits{J.}},
\oauthor{\bsnm{Subramanian}, \binits{S.}},
\oauthor{\bsnm{Harrington}, \binits{P.}},
\oauthor{\bsnm{Raja}, \binits{S.}},
\oauthor{\bsnm{Chattopadhyay}, \binits{A.}},
\oauthor{\bsnm{Mardani}, \binits{M.}},
\oauthor{\bsnm{Kurth}, \binits{T.}},
\oauthor{\bsnm{Hall}, \binits{D.}},
\oauthor{\bsnm{Li}, \binits{Z.}},
\oauthor{\bsnm{Azizzadenesheli}, \binits{K.}}, et al.:
Fourcastnet: A global data-driven high-resolution weather model using adaptive fourier neural operators.
arXiv preprint arXiv:2202.11214
(2022)
\end{botherref}
\endbibitem

\bibitem{bi2022pangu}
\begin{botherref}
\oauthor{\bsnm{Bi}, \binits{K.}},
\oauthor{\bsnm{Xie}, \binits{L.}},
\oauthor{\bsnm{Zhang}, \binits{H.}},
\oauthor{\bsnm{Chen}, \binits{X.}},
\oauthor{\bsnm{Gu}, \binits{X.}},
\oauthor{\bsnm{Tian}, \binits{Q.}}:
Pangu-weather: A 3d high-resolution model for fast and accurate global weather forecast.
arXiv preprint arXiv:2211.02556
(2022)
\end{botherref}
\endbibitem

\bibitem{lam2023learning}
\begin{barticle}
\bauthor{\bsnm{Lam}, \binits{R.}},
\bauthor{\bsnm{Sanchez-Gonzalez}, \binits{A.}},
\bauthor{\bsnm{Willson}, \binits{M.}},
\bauthor{\bsnm{Wirnsberger}, \binits{P.}},
\bauthor{\bsnm{Fortunato}, \binits{M.}},
\bauthor{\bsnm{Alet}, \binits{F.}},
\bauthor{\bsnm{Ravuri}, \binits{S.}},
\bauthor{\bsnm{Ewalds}, \binits{T.}},
\bauthor{\bsnm{Eaton-Rosen}, \binits{Z.}},
\bauthor{\bsnm{Hu}, \binits{W.}}, \betal:
\batitle{Learning skillful medium-range global weather forecasting}.
\bjtitle{Science}
\bvolume{382}(\bissue{6677}),
\bfpage{1416}--\blpage{1421}
(\byear{2023})
\end{barticle}
\endbibitem

\bibitem{chen2023fuxi}
\begin{barticle}
\bauthor{\bsnm{Chen}, \binits{L.}},
\bauthor{\bsnm{Zhong}, \binits{X.}},
\bauthor{\bsnm{Zhang}, \binits{F.}},
\bauthor{\bsnm{Cheng}, \binits{Y.}},
\bauthor{\bsnm{Xu}, \binits{Y.}},
\bauthor{\bsnm{Qi}, \binits{Y.}},
\bauthor{\bsnm{Li}, \binits{H.}}:
\batitle{Fuxi: a cascade machine learning forecasting system for 15-day global weather forecast}.
\bjtitle{npj climate and atmospheric science}
\bvolume{6}(\bissue{1}),
\bfpage{190}
(\byear{2023})
\end{barticle}
\endbibitem

\bibitem{lang2025multi}
\begin{botherref}
\oauthor{\bsnm{Lang}, \binits{S.}},
\oauthor{\bsnm{Leutbecher}, \binits{M.}},
\oauthor{\bsnm{Maciel}, \binits{P.}}:
A multi-scale loss formulation for learning a probabilistic model with proper score optimisation.
arXiv preprint arXiv:2506.10868
(2025)
\end{botherref}
\endbibitem

\bibitem{price2025probabilistic}
\begin{barticle}
\bauthor{\bsnm{Price}, \binits{I.}},
\bauthor{\bsnm{Sanchez-Gonzalez}, \binits{A.}},
\bauthor{\bsnm{Alet}, \binits{F.}},
\bauthor{\bsnm{Andersson}, \binits{T.R.}},
\bauthor{\bsnm{El-Kadi}, \binits{A.}},
\bauthor{\bsnm{Masters}, \binits{D.}},
\bauthor{\bsnm{Ewalds}, \binits{T.}},
\bauthor{\bsnm{Stott}, \binits{J.}},
\bauthor{\bsnm{Mohamed}, \binits{S.}},
\bauthor{\bsnm{Battaglia}, \binits{P.}}, \betal:
\batitle{Probabilistic weather forecasting with machine learning}.
\bjtitle{Nature}
\bvolume{637}(\bissue{8044}),
\bfpage{84}--\blpage{90}
(\byear{2025})
\end{barticle}
\endbibitem

\bibitem{zhong2024fuxi}
\begin{botherref}
\oauthor{\bsnm{Zhong}, \binits{X.}},
\oauthor{\bsnm{Chen}, \binits{L.}},
\oauthor{\bsnm{Li}, \binits{H.}},
\oauthor{\bsnm{Liu}, \binits{J.}},
\oauthor{\bsnm{Fan}, \binits{X.}},
\oauthor{\bsnm{Feng}, \binits{J.}},
\oauthor{\bsnm{Dai}, \binits{K.}},
\oauthor{\bsnm{Luo}, \binits{J.-J.}},
\oauthor{\bsnm{Wu}, \binits{J.}},
\oauthor{\bsnm{Lu}, \binits{B.}}:
Fuxi-ens: A machine learning model for medium-range ensemble weather forecasting.
arXiv preprint arXiv:2405.05925
(2024)
\end{botherref}
\endbibitem

\bibitem{bonavita2016evolution}
\begin{barticle}
\bauthor{\bsnm{Bonavita}, \binits{M.}},
\bauthor{\bsnm{H{\'o}lm}, \binits{E.}},
\bauthor{\bsnm{Isaksen}, \binits{L.}},
\bauthor{\bsnm{Fisher}, \binits{M.}}:
\batitle{The evolution of the ecmwf hybrid data assimilation system}.
\bjtitle{Quarterly Journal of the Royal Meteorological Society}
\bvolume{142}(\bissue{694}),
\bfpage{287}--\blpage{303}
(\byear{2016})
\end{barticle}
\endbibitem

\bibitem{houtekamer2016review}
\begin{barticle}
\bauthor{\bsnm{Houtekamer}, \binits{P.L.}},
\bauthor{\bsnm{Zhang}, \binits{F.}}:
\batitle{Review of the ensemble kalman filter for atmospheric data assimilation}.
\bjtitle{Monthly Weather Review}
\bvolume{144}(\bissue{12}),
\bfpage{4489}--\blpage{4532}
(\byear{2016})
\end{barticle}
\endbibitem

\bibitem{palmer1998singular}
\begin{barticle}
\bauthor{\bsnm{Palmer}, \binits{T.}},
\bauthor{\bsnm{Gelaro}, \binits{R.}},
\bauthor{\bsnm{Barkmeijer}, \binits{J.}},
\bauthor{\bsnm{Buizza}, \binits{R.}}:
\batitle{Singular vectors, metrics, and adaptive observations}.
\bjtitle{Journal of the Atmospheric Sciences}
\bvolume{55}(\bissue{4}),
\bfpage{633}--\blpage{653}
(\byear{1998})
\end{barticle}
\endbibitem

\bibitem{ehrendorfer1999singular}
\begin{barticle}
\bauthor{\bsnm{Ehrendorfer}, \binits{M.}},
\bauthor{\bsnm{Errico}, \binits{R.M.}},
\bauthor{\bsnm{Raeder}, \binits{K.D.}}:
\batitle{Singular-vector perturbation growth in a primitive equation model with moist physics}.
\bjtitle{Journal of the Atmospheric Sciences}
\bvolume{56}(\bissue{11}),
\bfpage{1627}--\blpage{1648}
(\byear{1999})
\end{barticle}
\endbibitem

\bibitem{gray1968global}
\begin{barticle}
\bauthor{\bsnm{Gray}, \binits{W.M.}}:
\batitle{Global view of the origin of tropical disturbances and storms}.
\bjtitle{Monthly Weather Review}
\bvolume{96}(\bissue{10}),
\bfpage{669}--\blpage{700}
(\byear{1968})
\end{barticle}
\endbibitem

\bibitem{emanuel1986air}
\begin{barticle}
\bauthor{\bsnm{Emanuel}, \binits{K.A.}}:
\batitle{An air-sea interaction theory for tropical cyclones. part i: Steady-state maintenance}.
\bjtitle{Journal of Atmospheric Sciences}
\bvolume{43}(\bissue{6}),
\bfpage{585}--\blpage{605}
(\byear{1986})
\end{barticle}
\endbibitem

\bibitem{done2022response}
\begin{barticle}
\bauthor{\bsnm{Done}, \binits{J.M.}},
\bauthor{\bsnm{Lackmann}, \binits{G.M.}},
\bauthor{\bsnm{Prein}, \binits{A.F.}}:
\batitle{The response of tropical cyclone intensity to changes in environmental temperature}.
\bjtitle{Weather and Climate Dynamics}
\bvolume{3}(\bissue{2}),
\bfpage{693}--\blpage{711}
(\byear{2022})
\end{barticle}
\endbibitem

\bibitem{richardson2000skill}
\begin{barticle}
\bauthor{\bsnm{Richardson}, \binits{D.S.}}:
\batitle{Skill and relative economic value of the ecmwf ensemble prediction system}.
\bjtitle{Quarterly Journal of the Royal Meteorological Society}
\bvolume{126}(\bissue{563}),
\bfpage{649}--\blpage{667}
(\byear{2000})
\end{barticle}
\endbibitem

\bibitem{dulac2024assessing}
\begin{barticle}
\bauthor{\bsnm{Dulac}, \binits{W.}},
\bauthor{\bsnm{Cattiaux}, \binits{J.}},
\bauthor{\bsnm{Chauvin}, \binits{F.}},
\bauthor{\bsnm{Bourdin}, \binits{S.}},
\bauthor{\bsnm{Fromang}, \binits{S.}}:
\batitle{Assessing the representation of tropical cyclones in era5 with the cnrm tracker}.
\bjtitle{Climate Dynamics}
\bvolume{62}(\bissue{1}),
\bfpage{223}--\blpage{238}
(\byear{2024})
\end{barticle}
\endbibitem

\bibitem{liu2024evaluation}
\begin{barticle}
\bauthor{\bsnm{Liu}, \binits{C.-C.}},
\bauthor{\bsnm{Hsu}, \binits{K.}},
\bauthor{\bsnm{Peng}, \binits{M.S.}},
\bauthor{\bsnm{Chen}, \binits{D.-S.}},
\bauthor{\bsnm{Chang}, \binits{P.-L.}},
\bauthor{\bsnm{Hsiao}, \binits{L.-F.}},
\bauthor{\bsnm{Fong}, \binits{C.-T.}},
\bauthor{\bsnm{Hong}, \binits{J.-S.}},
\bauthor{\bsnm{Cheng}, \binits{C.-P.}},
\bauthor{\bsnm{Lu}, \binits{K.-C.}}, \betal:
\batitle{Evaluation of five global ai models for predicting weather in eastern asia and western pacific}.
\bjtitle{npj Climate and Atmospheric Science}
\bvolume{7}(\bissue{1}),
\bfpage{221}
(\byear{2024})
\end{barticle}
\endbibitem

\bibitem{subich2025fixing}
\begin{botherref}
\oauthor{\bsnm{Subich}, \binits{C.}},
\oauthor{\bsnm{Husain}, \binits{S.Z.}},
\oauthor{\bsnm{Separovic}, \binits{L.}},
\oauthor{\bsnm{Yang}, \binits{J.}}:
Fixing the double penalty in data-driven weather forecasting through a modified spherical harmonic loss function.
arXiv preprint arXiv:2501.19374
(2025)
\end{botherref}
\endbibitem

\bibitem{hersbach2020era5}
\begin{barticle}
\bauthor{\bsnm{Hersbach}, \binits{H.}},
\bauthor{\bsnm{Bell}, \binits{B.}},
\bauthor{\bsnm{Berrisford}, \binits{P.}},
\bauthor{\bsnm{Hirahara}, \binits{S.}},
\bauthor{\bsnm{Hor{\'a}nyi}, \binits{A.}},
\bauthor{\bsnm{Mu{\~n}oz-Sabater}, \binits{J.}},
\bauthor{\bsnm{Nicolas}, \binits{J.}},
\bauthor{\bsnm{Peubey}, \binits{C.}},
\bauthor{\bsnm{Radu}, \binits{R.}},
\bauthor{\bsnm{Schepers}, \binits{D.}}, \betal:
\batitle{The era5 global reanalysis}.
\bjtitle{Quarterly journal of the royal meteorological society}
\bvolume{146}(\bissue{730}),
\bfpage{1999}--\blpage{2049}
(\byear{2020})
\end{barticle}
\endbibitem

\bibitem{knapp2010international}
\begin{barticle}
\bauthor{\bsnm{Knapp}, \binits{K.R.}},
\bauthor{\bsnm{Kruk}, \binits{M.C.}},
\bauthor{\bsnm{Levinson}, \binits{D.H.}},
\bauthor{\bsnm{Diamond}, \binits{H.J.}},
\bauthor{\bsnm{Neumann}, \binits{C.J.}}:
\batitle{The international best track archive for climate stewardship (ibtracs) unifying tropical cyclone data}.
\bjtitle{Bulletin of the American Meteorological Society}
\bvolume{91}(\bissue{3}),
\bfpage{363}--\blpage{376}
(\byear{2010})
\end{barticle}
\endbibitem

\bibitem{kenneth2019international}
\begin{botherref}
\oauthor{\bsnm{Kenneth}, \binits{R.}},
\oauthor{\bsnm{Howard}, \binits{J.}},
\oauthor{\bsnm{James}, \binits{P.}},
\oauthor{\bsnm{Michael}, \binits{C.}},
\oauthor{\bsnm{Carl}, \binits{J.}}:
International best track archive for climate stewardship (ibtracs) project, version 4.
(No Title)
(2019)
\end{botherref}
\endbibitem

\bibitem{kingma2013auto}
\begin{botherref}
\oauthor{\bsnm{Kingma}, \binits{D.P.}},
\oauthor{\bsnm{Welling}, \binits{M.}}:
Auto-encoding variational bayes.
arXiv preprint arXiv:1312.6114
(2013)
\end{botherref}
\endbibitem

\bibitem{kingma2019introduction}
\begin{barticle}
\bauthor{\bsnm{Kingma}, \binits{D.P.}},
\bauthor{\bsnm{Welling}, \binits{M.}}, \betal:
\batitle{An introduction to variational autoencoders}.
\bjtitle{Foundations and Trends{\textregistered} in Machine Learning}
\bvolume{12}(\bissue{4}),
\bfpage{307}--\blpage{392}
(\byear{2019})
\end{barticle}
\endbibitem

\bibitem{Liu2021Swin}
\begin{botherref}
\oauthor{\bsnm{Liu}, \binits{Z.}},
\oauthor{\bsnm{Lin}, \binits{Y.}},
\oauthor{\bsnm{Cao}, \binits{Y.}},
\oauthor{\bsnm{Hu}, \binits{H.}},
\oauthor{\bsnm{Wei}, \binits{Y.}},
\oauthor{\bsnm{Zhang}, \binits{Z.}},
\oauthor{\bsnm{Lin}, \binits{S.}},
\oauthor{\bsnm{Guo}, \binits{B.}}:
Swin transformer: Hierarchical vision transformer using shifted windows.
2021 IEEE/CVF International Conference on Computer Vision (ICCV),
9992--10002
(2021).
\doiurl{10.1109/iccv48922.2021.00986}
\end{botherref}
\endbibitem

\bibitem{zhong2024fuxiextreme}
\begin{barticle}
\bauthor{\bsnm{Zhong}, \binits{X.}},
\bauthor{\bsnm{Chen}, \binits{L.}},
\bauthor{\bsnm{Liu}, \binits{J.}},
\bauthor{\bsnm{Lin}, \binits{C.}},
\bauthor{\bsnm{Qi}, \binits{Y.}},
\bauthor{\bsnm{Li}, \binits{H.}}:
\batitle{Fuxi-extreme: Improving extreme rainfall and wind forecasts with diffusion model}.
\bjtitle{Science China Earth Sciences}
\bvolume{67}(\bissue{12}),
\bfpage{3696}--\blpage{3708}
(\byear{2024})
\end{barticle}
\endbibitem

\bibitem{zhang2015verification}
\begin{barticle}
\bauthor{\bsnm{Zhang}, \binits{X.}},
\bauthor{\bsnm{Chen}, \binits{G.}},
\bauthor{\bsnm{Yu}, \binits{H.}},
\bauthor{\bsnm{Zeng}, \binits{Z.}}:
\batitle{Verification of ensemble track forecasts of tropical cyclones during 2014}.
\bjtitle{Tropical Cyclone Research and Review}
\bvolume{4}(\bissue{2}),
\bfpage{79}--\blpage{87}
(\byear{2015})
\end{barticle}
\endbibitem

\bibitem{mason2002areas}
\begin{barticle}
\bauthor{\bsnm{Mason}, \binits{S.J.}},
\bauthor{\bsnm{Graham}, \binits{N.E.}}:
\batitle{Areas beneath the relative operating characteristics (roc) and relative operating levels (rol) curves: Statistical significance and interpretation}.
\bjtitle{Quarterly Journal of the Royal Meteorological Society: A journal of the atmospheric sciences, applied meteorology and physical oceanography}
\bvolume{128}(\bissue{584}),
\bfpage{2145}--\blpage{2166}
(\byear{2002})
\end{barticle}
\endbibitem

\bibitem{TheRelationshipbetweenEnsembleSpreadandEnsembleMeanSkill}
\begin{barticle}
\bauthor{\bsnm{Whitaker}, \binits{J.S.}},
\bauthor{\bsnm{Loughe}, \binits{A.F.}}:
\batitle{The relationship between ensemble spread and ensemble mean skill}.
\bjtitle{Monthly Weather Review}
\bvolume{126}(\bissue{12}),
\bfpage{3292}--\blpage{3302}
(\byear{1998})
\end{barticle}
\endbibitem

\bibitem{yamaguchi2009typhoon}
\begin{barticle}
\bauthor{\bsnm{Yamaguchi}, \binits{M.}},
\bauthor{\bsnm{Sakai}, \binits{R.}},
\bauthor{\bsnm{Kyoda}, \binits{M.}},
\bauthor{\bsnm{Komori}, \binits{T.}},
\bauthor{\bsnm{Kadowaki}, \binits{T.}}:
\batitle{Typhoon ensemble prediction system developed at the japan meteorological agency}.
\bjtitle{Monthly Weather Review}
\bvolume{137}(\bissue{8}),
\bfpage{2592}--\blpage{2604}
(\byear{2009})
\end{barticle}
\endbibitem

\bibitem{goerss2000tropical}
\begin{barticle}
\bauthor{\bsnm{Goerss}, \binits{J.S.}}:
\batitle{Tropical cyclone track forecasts using an ensemble of dynamical models}.
\bjtitle{Monthly Weather Review}
\bvolume{128}(\bissue{4}),
\bfpage{1187}--\blpage{1193}
(\byear{2000})
\end{barticle}
\endbibitem

\bibitem{aberson2001ensemble}
\begin{barticle}
\bauthor{\bsnm{Aberson}, \binits{S.D.}}:
\batitle{The ensemble of tropical cyclone track forecasting models in the north atlantic basin (1976--2000)}.
\bjtitle{Bulletin of the American Meteorological Society}
\bvolume{82}(\bissue{9}),
\bfpage{1895}--\blpage{1904}
(\byear{2001})
\end{barticle}
\endbibitem

\bibitem{cangialosi2012national}
\begin{botherref}
\oauthor{\bsnm{Cangialosi}, \binits{J.P.}},
\oauthor{\bsnm{Franklin}, \binits{J.}}:
National hurricane center forecast verification report.
National Hurricane Center (NHC)
\textbf{79}
(2012)
\end{botherref}
\endbibitem

\bibitem{goerss2004history}
\begin{barticle}
\bauthor{\bsnm{Goerss}, \binits{J.S.}},
\bauthor{\bsnm{Sampson}, \binits{C.R.}},
\bauthor{\bsnm{Gross}, \binits{J.M.}}:
\batitle{A history of western north pacific tropical cyclone track forecast skill}.
\bjtitle{Weather and Forecasting}
\bvolume{19}(\bissue{3}),
\bfpage{633}--\blpage{638}
(\byear{2004})
\end{barticle}
\endbibitem

\bibitem{kirkland2019regional}
\begin{barticle}
\bauthor{\bsnm{Kirkland}, \binits{J.L.}},
\bauthor{\bsnm{Zick}, \binits{S.E.}}:
\batitle{Regional differences in the spatial patterns of north atlantic tropical cyclone rainbands through landfall}.
\bjtitle{southeastern geographer}
\bvolume{59}(\bissue{3}),
\bfpage{294}--\blpage{320}
(\byear{2019})
\end{barticle}
\endbibitem

\bibitem{nachar2008mann}
\begin{barticle}
\bauthor{\bsnm{Nachar}, \binits{N.}}, \betal:
\batitle{The mann-whitney u: A test for assessing whether two independent samples come from the same distribution}.
\bjtitle{Tutorials in quantitative Methods for Psychology}
\bvolume{4}(\bissue{1}),
\bfpage{13}--\blpage{20}
(\byear{2008})
\end{barticle}
\endbibitem

\bibitem{strikeprobability}
\begin{barticle}
\bauthor{\bsnm{Titley}, \binits{H.A.}},
\bauthor{\bsnm{Bowyer}, \binits{R.L.}},
\bauthor{\bsnm{Cloke}, \binits{H.L.}}:
\batitle{A global evaluation of multi-model ensemble tropical cyclone track probability forecasts}.
\bjtitle{Quarterly Journal of the Royal Meteorological Society}
\bvolume{146}(\bissue{726}),
\bfpage{531}--\blpage{545}
(\byear{2020}).
\doiurl{10.1002/qj.3712}
\end{barticle}
\endbibitem

\bibitem{strikeprobability2}
\begin{barticle}
\bauthor{\bsnm{Gentile}, \binits{E.S.}},
\bauthor{\bsnm{Gray}, \binits{S.L.}},
\bauthor{\bsnm{Lewis}, \binits{H.W.}}:
\batitle{The sensitivity of probabilistic convective-scale forecasts of an extratropical cyclone to atmosphere–ocean–wave coupling}.
\bjtitle{Quarterly Journal of the Royal Meteorological Society}
\bvolume{148}(\bissue{743}),
\bfpage{685}--\blpage{710}
(\byear{2022}).
\doiurl{10.1002/qj.4225}
\end{barticle}
\endbibitem

\bibitem{shao2024impact}
\begin{barticle}
\bauthor{\bsnm{Shao}, \binits{D.}},
\bauthor{\bsnm{Zhang}, \binits{Y.}},
\bauthor{\bsnm{Xu}, \binits{J.}}, \betal:
\batitle{Impact of model perturbation schemes on regional ensemble forecasting of typhoons based on typhoon kompasu (2118)}.
\bjtitle{Chinese Journal of Atmospheric Sciences}
\bvolume{48}(\bissue{4}),
\bfpage{1511}--\blpage{1530}
(\byear{2024}).
\doiurl{10.3878/j.issn.1006-9895.2301.22122}
\end{barticle}
\endbibitem

\end{thebibliography}


\end{CJK*}

\clearpage

\pagestyle{plain} 
\appendix

\FloatBarrier

\setcounter{figure}{0}
\renewcommand{\figurename}{Supplementary Fig.}

\section*{Supplementary information}
\maketitle

\section*{Contents of this file}
\begin{itemize}
  \item[] Supplementary Table 1
  \item[] Supplementary Figures 1 to 11
\end{itemize}
\section*{Supplementary Table}


\begin{table}[h]
\centering
\caption{\label{summary} Parameters and thresholds related to the TC tracking.}
\begin{tabularx}{\textwidth}{cXXXX}
\hline
\textbf{Parameter} & \textbf{Level} & \textbf{Radius}  & \textbf{Threshold} & \textbf{Comments} \\
\hline
WS10M & Surface & 278 km & $>$ 8 m s$^{-1}$ & Only required over land. \\
Vorticity & 850 hPa & 278 km & absolute value $\geq 5 \times 10^{-5}$ s$^{-1}$ \\
\begin{tabular}{@{}c@{}}Geopotential \\ thickness\end{tabular} & 850 hPa and 200 hPa & 278 km & & A maximum in thickness is necessary only after the TC has transitioned into an extra-tropical cyclone. \\
\hline
\end{tabularx}
\end{table}

\section*{Supplementary Figures}
\begin{figure}[!htbp]
    \centering
    \includegraphics[width=\linewidth]{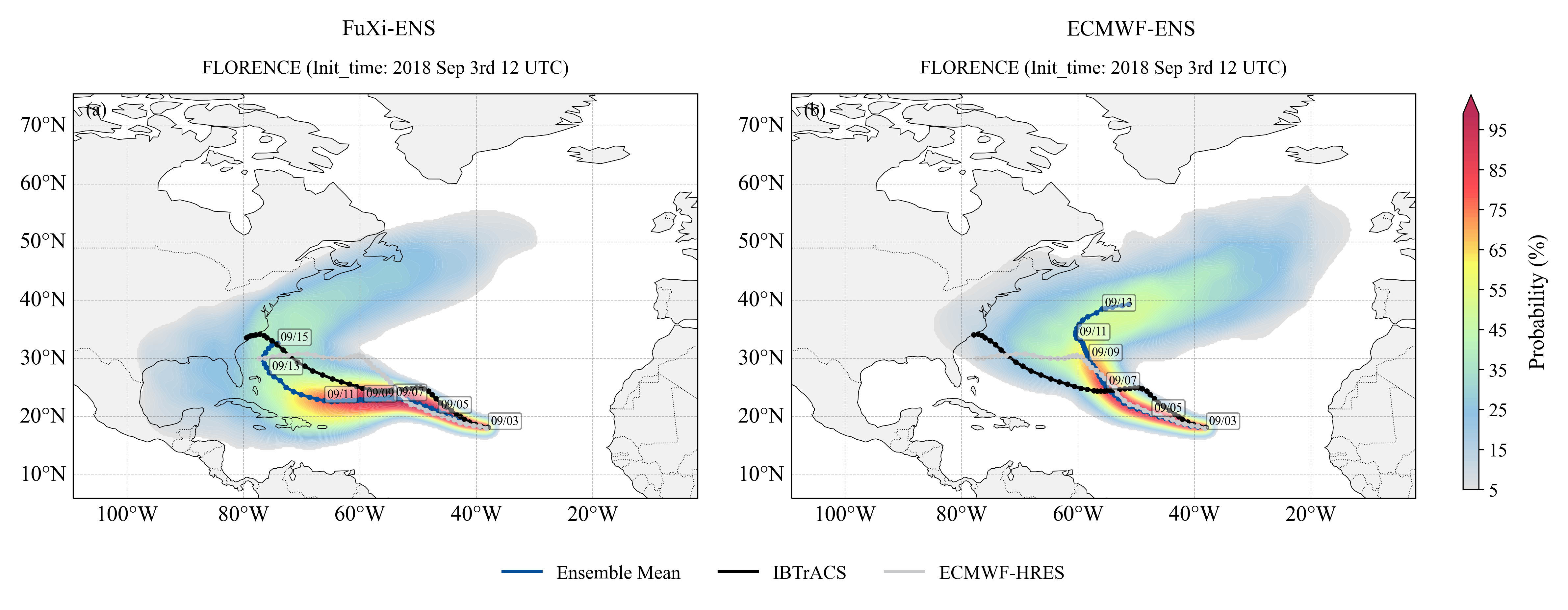}
    \caption{\textbf{Probabilistic forecast comparison for hurricane FLORENCE track prediction.} Hurricane FLORENCE track probabilistic forecast distribution based on initialization at 12 UTC on September 3, 2018: (\textbf{a}) FuXi-ENS and (\textbf{b}) ECMWF-ENS. Blue lines represent ensemble mean tracks, black lines represent IBTrACS observed tracks, and gray lines represent ECMWF-HRES (deterministic forecast). Shading indicates TC strike probability (\%). Date labels mark the temporal evolution of the track.}
    \label{model}    
\end{figure}

\begin{figure}[!htbp]
    \centering
    \includegraphics[width=\linewidth]{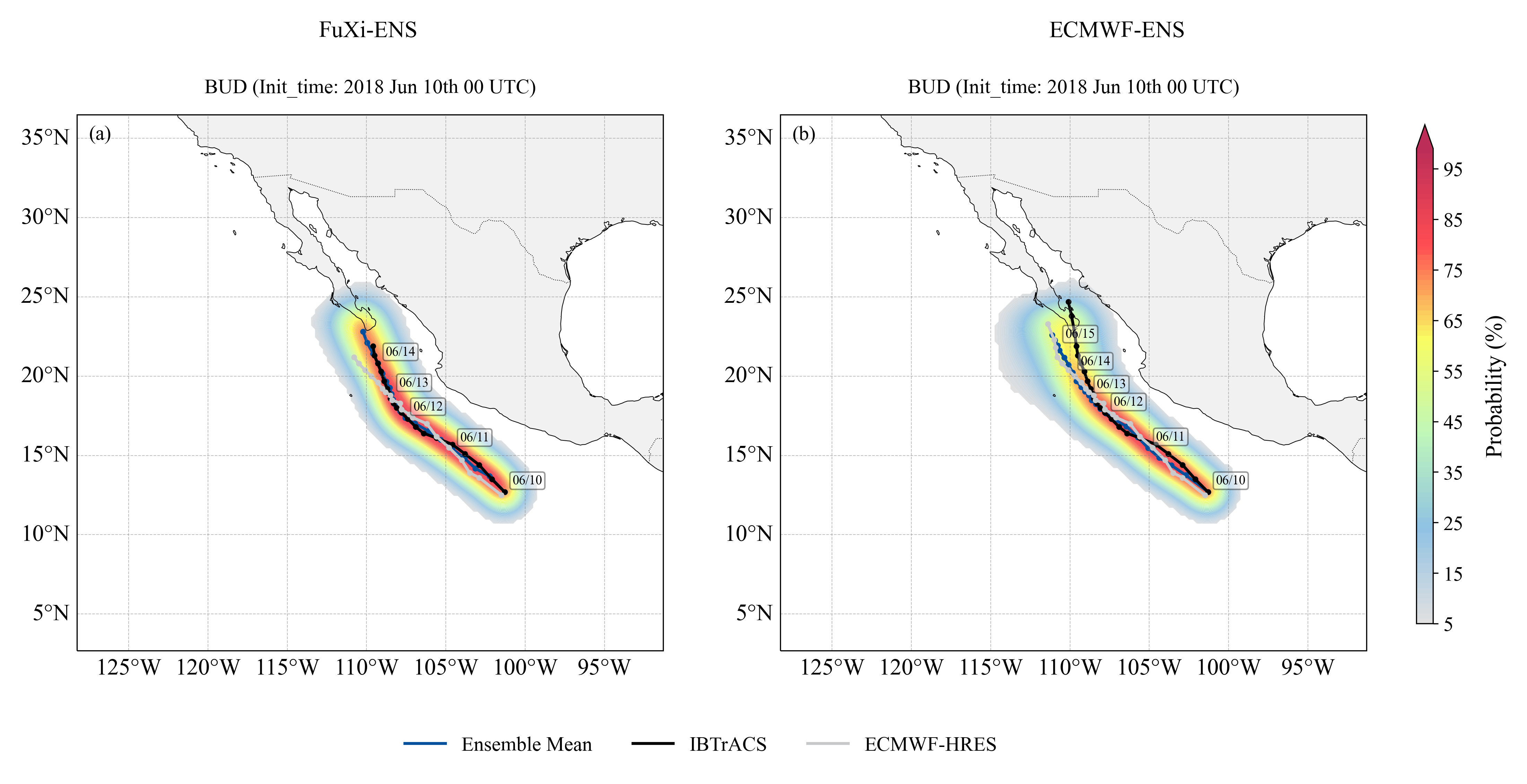}
    \caption{\textbf{Probabilistic forecast comparison for hurricane BUD track prediction.} Hurricane BUD track probabilistic forecast distribution based on initialization at 00 UTC on June 10, 2018: (\textbf{a}) FuXi-ENS and (\textbf{b}) ECMWF-ENS. Blue lines represent ensemble mean tracks, black lines represent IBTrACS observed tracks, and gray lines represent ECMWF-HRES (deterministic forecast). Shading indicates TC strike probability (\%). Date labels mark the temporal evolution of the track.}
    \label{model}    
\end{figure}

\begin{figure}[!htbp]
    \centering
    \includegraphics[width=\linewidth]{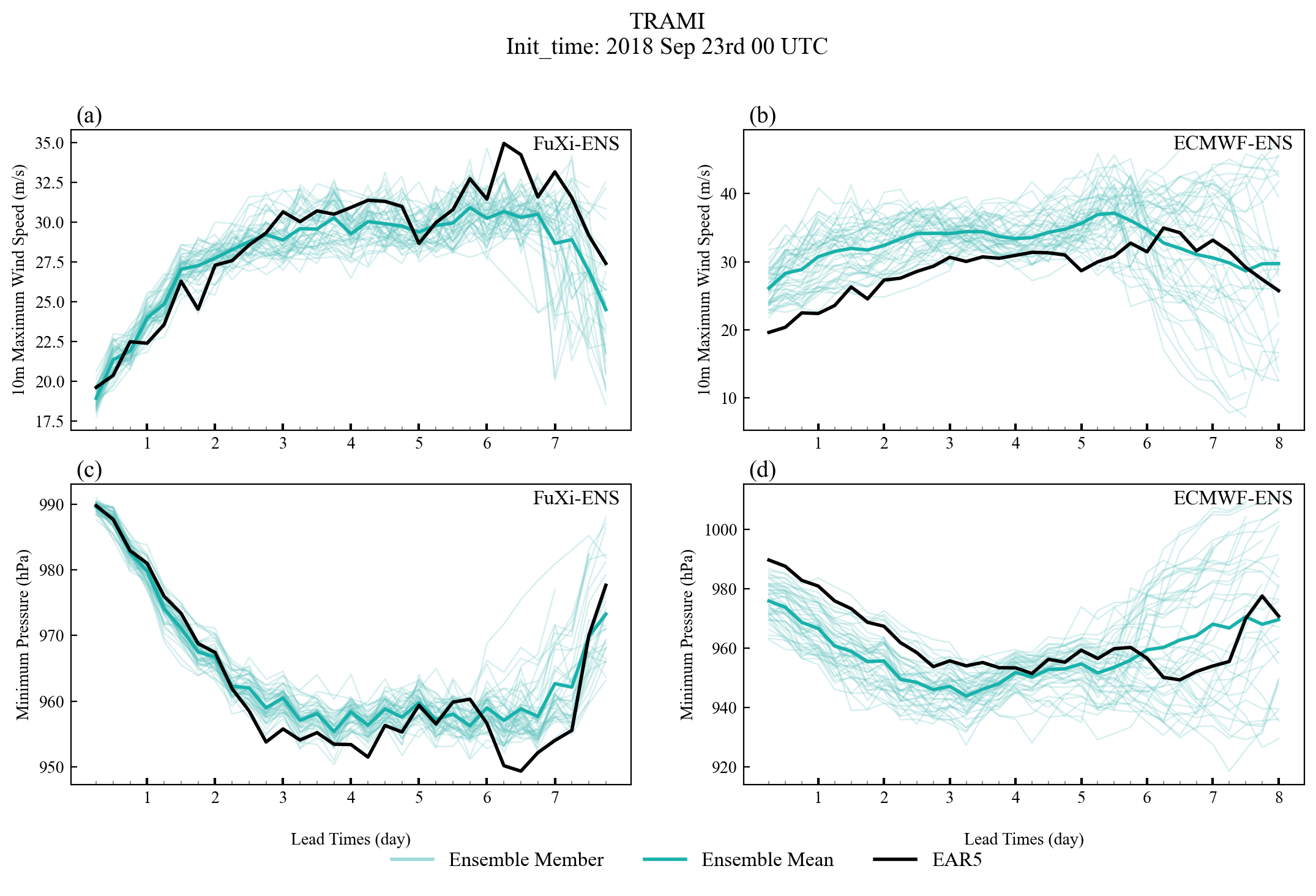}
    \caption{\textbf{Intensity forecast comparison for typhoon TRAMI.}
    10m maximum wind speed variation with forecast lead time for (\textbf{a}) FuXi-ENS and (\textbf{b}) ECMWF-ENS initialized at 00 UTC on September 23, 2018. Minimum pressure variation with forecast lead time for (\textbf{c}) FuXi-ENS and (\textbf{d}) ECMWF-ENS. Light green thin lines represent ensemble member forecasts, dark green thick lines represent ensemble mean forecasts, and black thick lines represent ERA5 reanalysis.}
    \label{model}    
\end{figure}

\begin{figure}[!htbp]
    \centering
    \includegraphics[width=\linewidth]{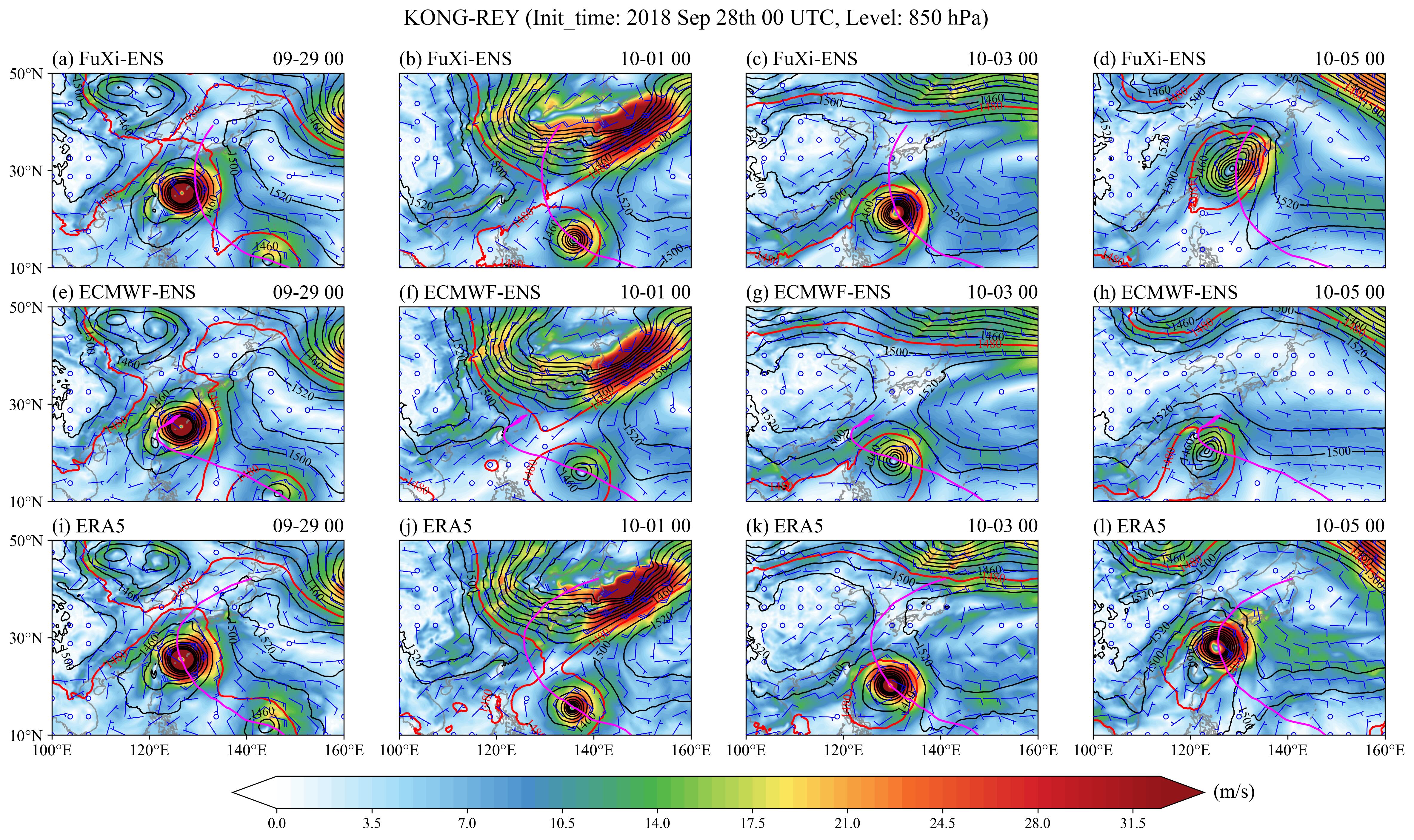}
    \caption{\textbf{Evolution of 850 hPa atmospheric circulation patterns for typhoon KONG-REY with forecast lead time.}
    Evolution of KONG-REY's 850 hPa wind speed (shading, m/s), geopotential height (contours, gpm, 20 gpm intervals), and horizontal wind (vectors, m/s) initialized at 00 UTC on September 28, 2018, for September 29 (first column), October 1 (second column), October 3 (third column), and October 5 (fourth column) from FuXi-ENS (first row), ECMWF-ENS (second row), and ERA5 (third row). Red contours indicate the 1480 gpm contour line, and magenta lines denote the best track (IBTrACS).}
    \label{model}    
\end{figure}

\begin{figure}[!htbp]
    \centering
    \includegraphics[width=\linewidth]{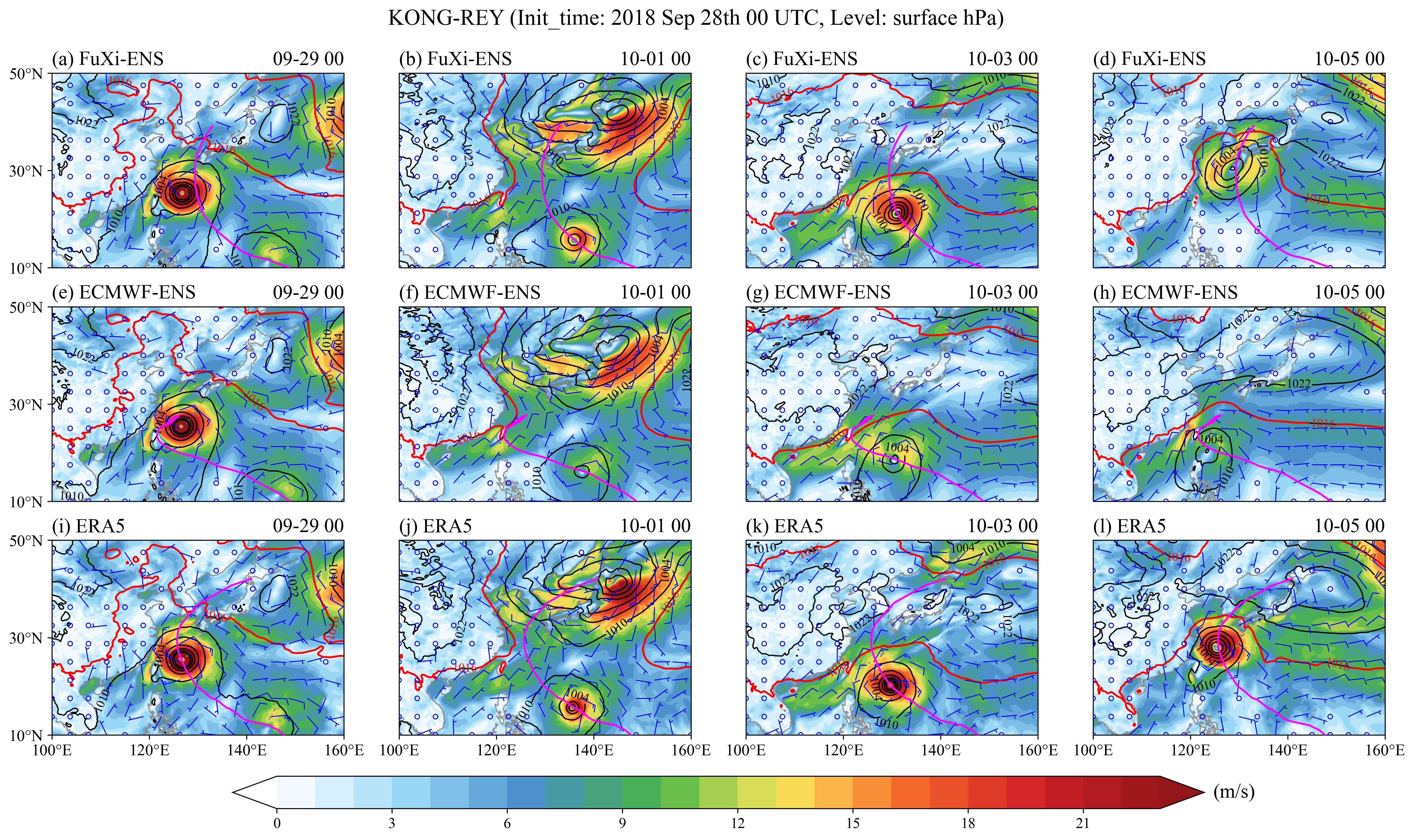}
    \caption{\textbf{Evolution of surface atmospheric circulation patterns for typhoon KONG-REY with forecast lead time.}
    Evolution of KONG-REY's 10m wind speed (shading, m/s), mean sea level pressure (contours, hPa, 6 hPa intervals), and 10m wind (vectors, m/s) initialized at 00 UTC on September 28, 2018, for September 29 (first column), October 1 (second column), October 3 (third column), and October 5 (fourth column) from FuXi-ENS (first row), ECMWF-ENS (second row), and ERA5 (third row). Red contours indicate the 1016 isoline, and magenta lines denote the best track (IBTrACS).}
    \label{model}    
\end{figure}

\begin{figure}[!htbp]
    \centering
    \includegraphics[width=\linewidth]{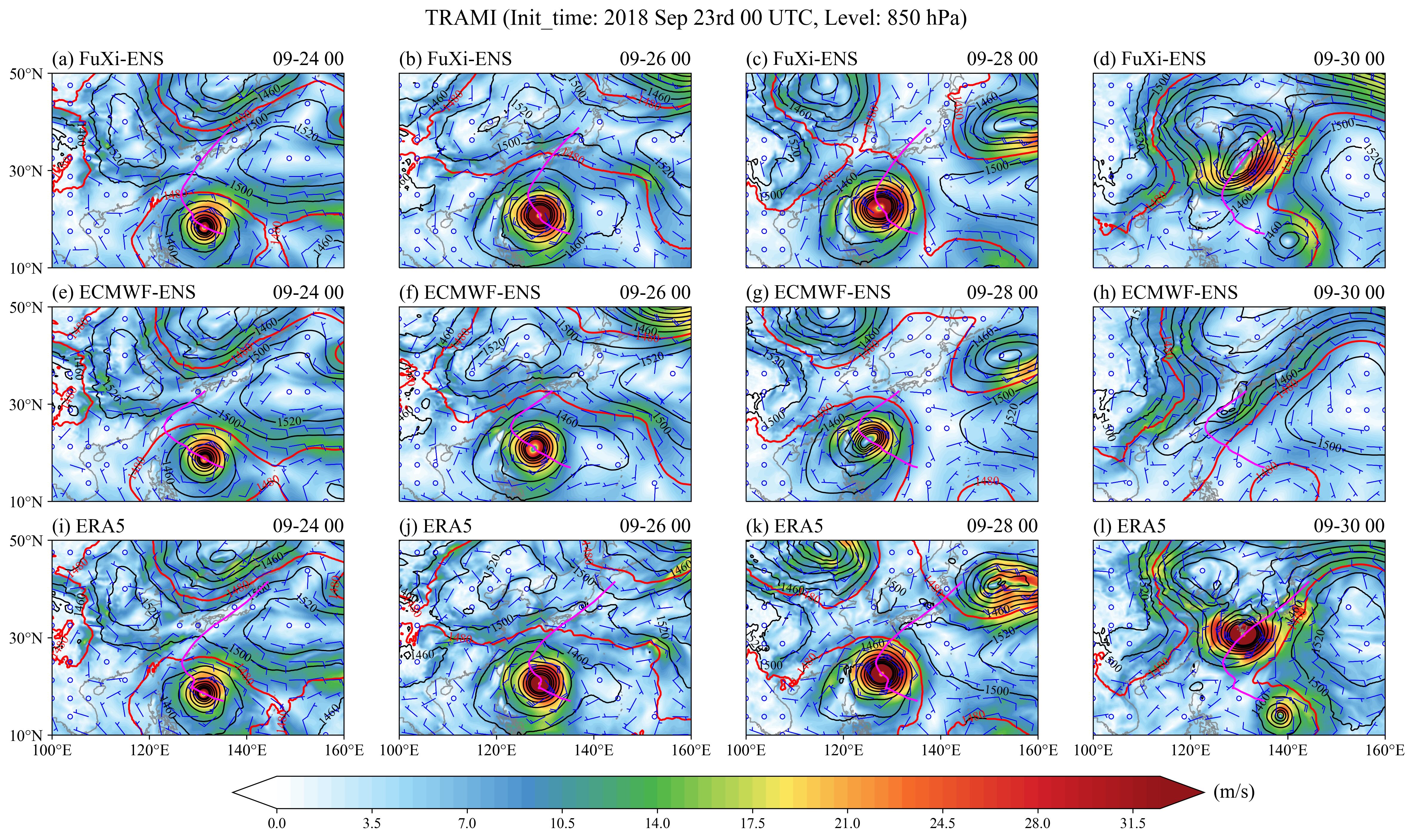}
    \caption{\textbf{Evolution of 850 hPa atmospheric circulation patterns for typhoon TRAMI with forecast lead time.}
    Evolution of TRAMI's 850 hPa wind speed (shading, m/s), geopotential height (contours, gpm, 20 gpm intervals), and horizontal wind (vectors, m/s) initialized at 00 UTC on September 23, 2018, for September 24 (first column), September 26 (second column), September 28 (third column), and September 30 (fourth column) from FuXi-ENS (first row), ECMWF-ENS (second row), and ERA5 (third row). Red contours indicate the 1480 gpm contour line, and magenta lines denote the best track (IBTrACS).}
    \label{model}    
\end{figure}

\begin{figure}[!htbp]
    \centering
    \includegraphics[width=\linewidth]{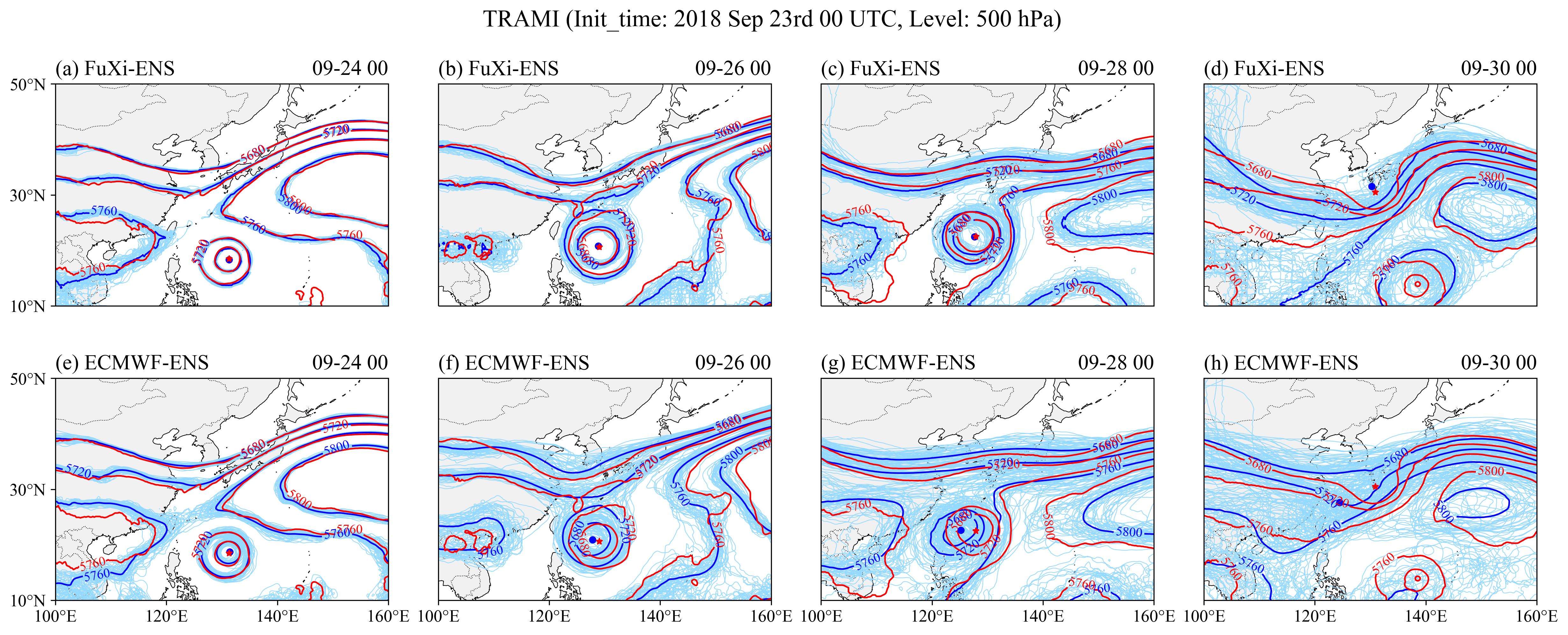}
    \caption{\textbf{Evolution of 500 hPa geopotential height ensemble forecasts for typhoon TRAMI with forecast lead time.}\\
    Evolution of TRAMI's 500 hPa geopotential height (contours, gpm) initialized at 00 UTC on September 23, 2018, for September 24 (first column), September 26 (second column), September 28 (third column), and September 30 (fourth column) from FuXi-ENS (first row), ECMWF-ENS (second row). Light blue lines represent individual ensemble members, dark blue lines represent the ensemble mean, and red lines represent ERA5 reanalysis. Four contour lines are shown: 5680 gpm, 5720 gpm, 5760 gpm and 5800 gpm. Blue dots denote the ensemble mean track and red stars denote the best track (IBTrACS).}
    \label{model}    
\end{figure}

\begin{figure}[!htbp]
    \centering
    \includegraphics[width=\linewidth]{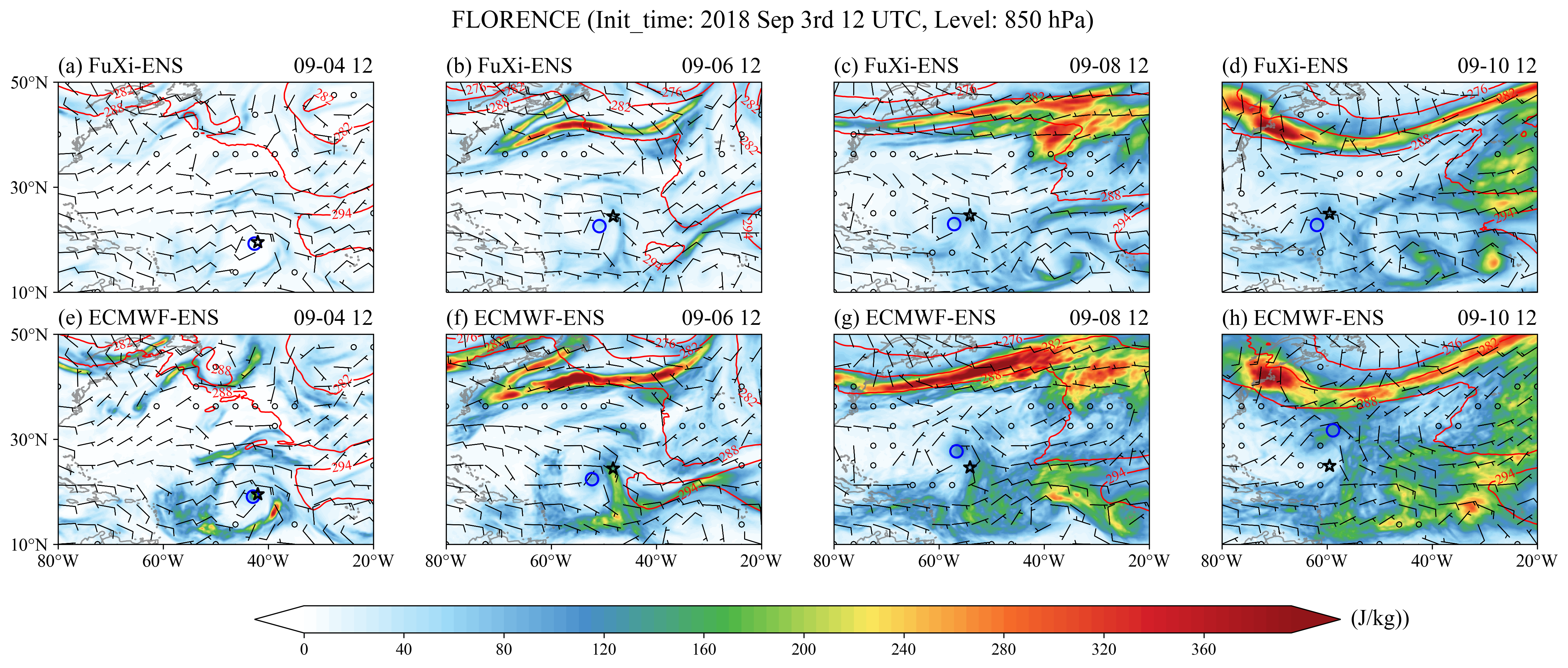}
    \caption{\textbf{Evolution of 850 hPa MTE for hurricane FLORENCE with forecast lead time.}
    Evolution of FLORENCE's 850 hPa MTE (shading, J/kg), temperature (contours, K), and horizontal wind (vectors, m/s) initialized at 12 UTC on September 3, 2018, for September 4 (first column), September 6 (second column), September 8 (third column), and September 10 (fourth column) from FuXi-ENS (first row), ECMWF-ENS (second row). Blue circles denote the ensemble mean track and black stars denote the best track (IBTrACS).}
    \label{model}    
\end{figure}

\begin{figure}[!htbp]
    \centering
    \includegraphics[width=\linewidth]{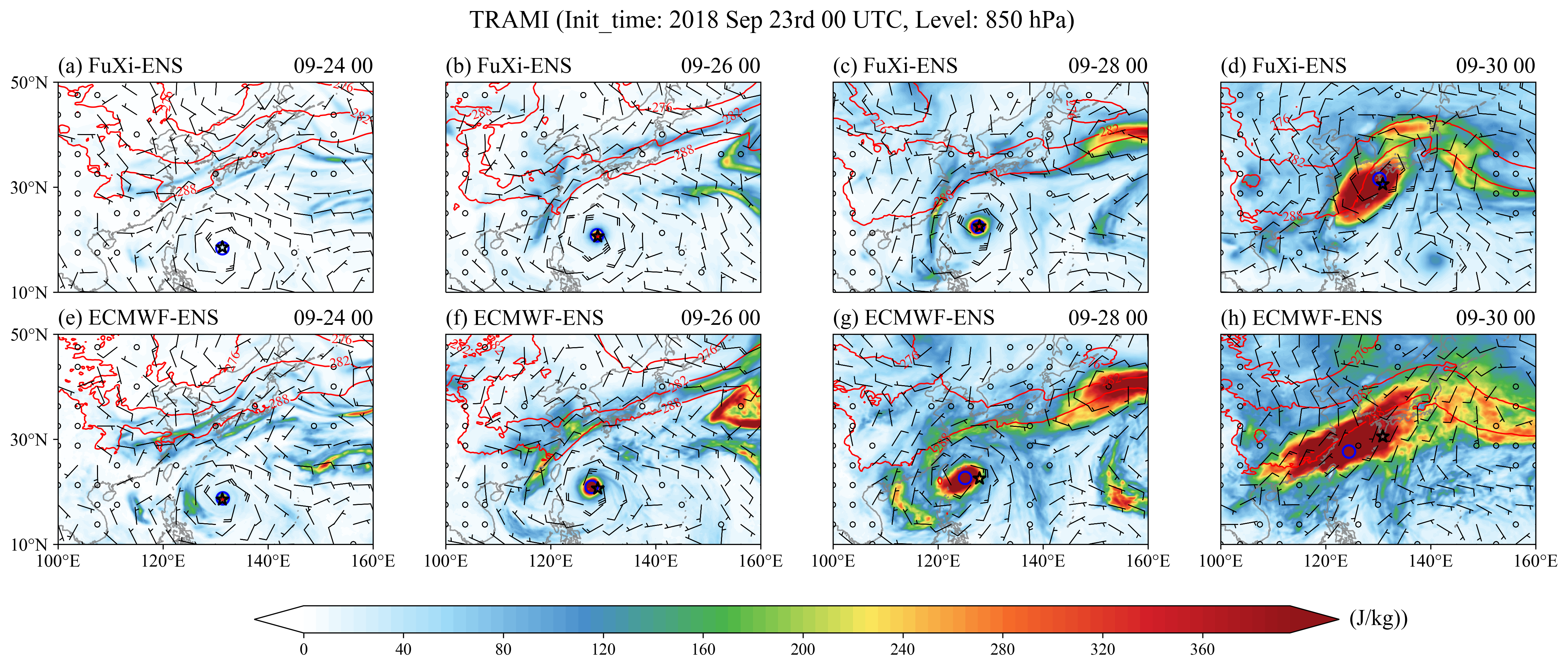}
    \caption{\textbf{Evolution of 850 hPa MTE for typhoon TRAMI with forecast lead time.}
    Evolution of TRAMI's 850 hPa MTE (shading, J/kg), temperature (contours, K), and horizontal wind (vectors, m/s) initialized at 00 UTC on September 28, 2018, for September 29 (first column), October 1 (second column), October 3 (third column), and October 5 (fourth column) from FuXi-ENS (first row), ECMWF-ENS (second row). Blue circles denote the ensemble mean track and black stars denote the best track (IBTrACS).}
    \label{model}    
\end{figure}

\begin{figure}[!htbp]
    \centering
    \includegraphics[width=0.8\linewidth]{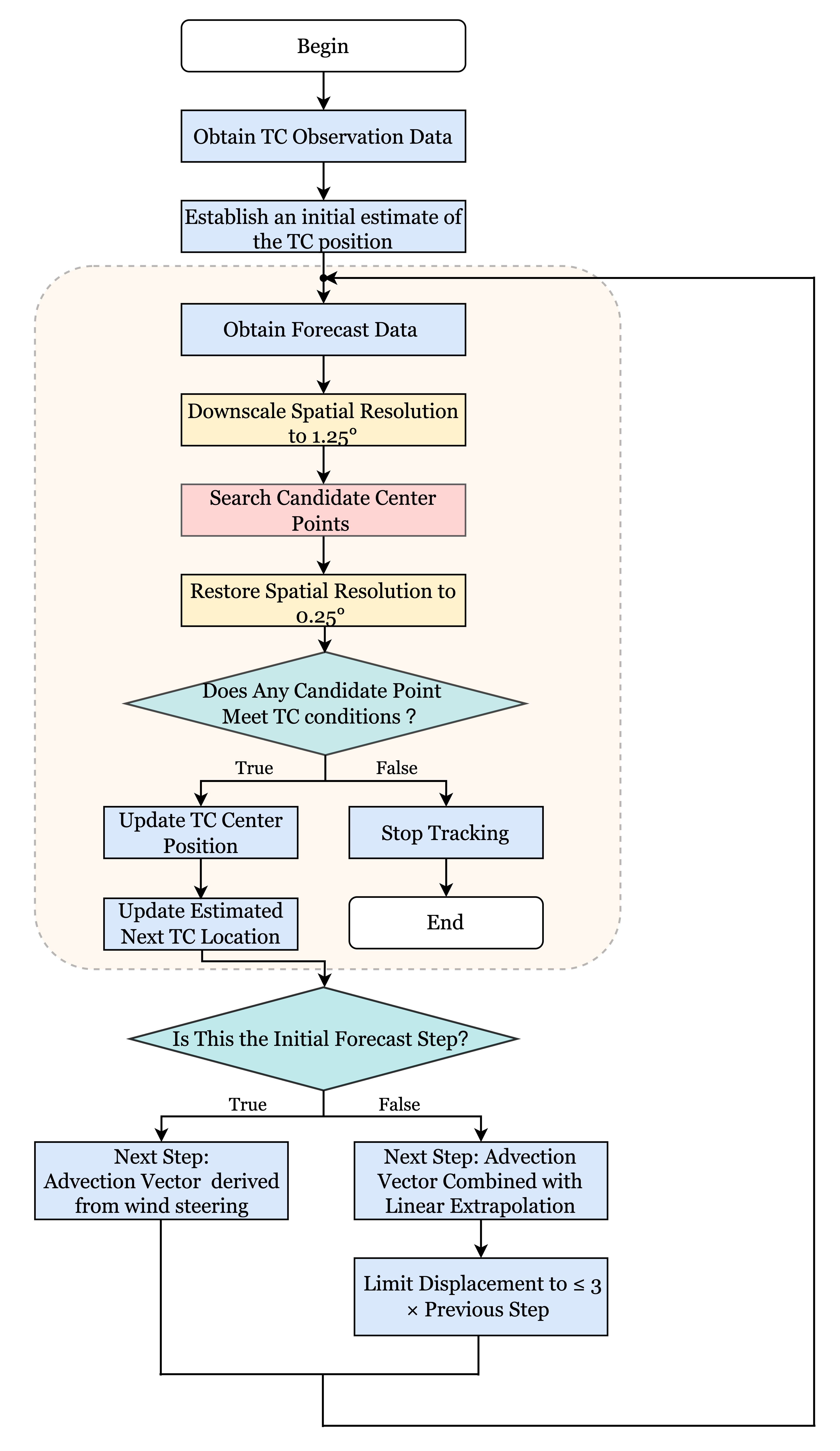}
    \caption{\textbf{Schematic diagram of the TC ensemble forecast tracking algorithm.} The algorithm obtains TC observation data to establish an initial estimate of the TC position, then obtains forecast data for downscaling spatial resolution and searching candidate center points. TC center position is updated when any candidate point meets TC conditions, otherwise stop tracking. Subsequent steps use advection vector derived from wind steering or advection vector combined with linear extrapolation, depending on whether this is the initial forecast step.}
    \label{model}    
\end{figure}

\begin{figure}[!htbp]
    \centering
    \includegraphics[width=\linewidth]{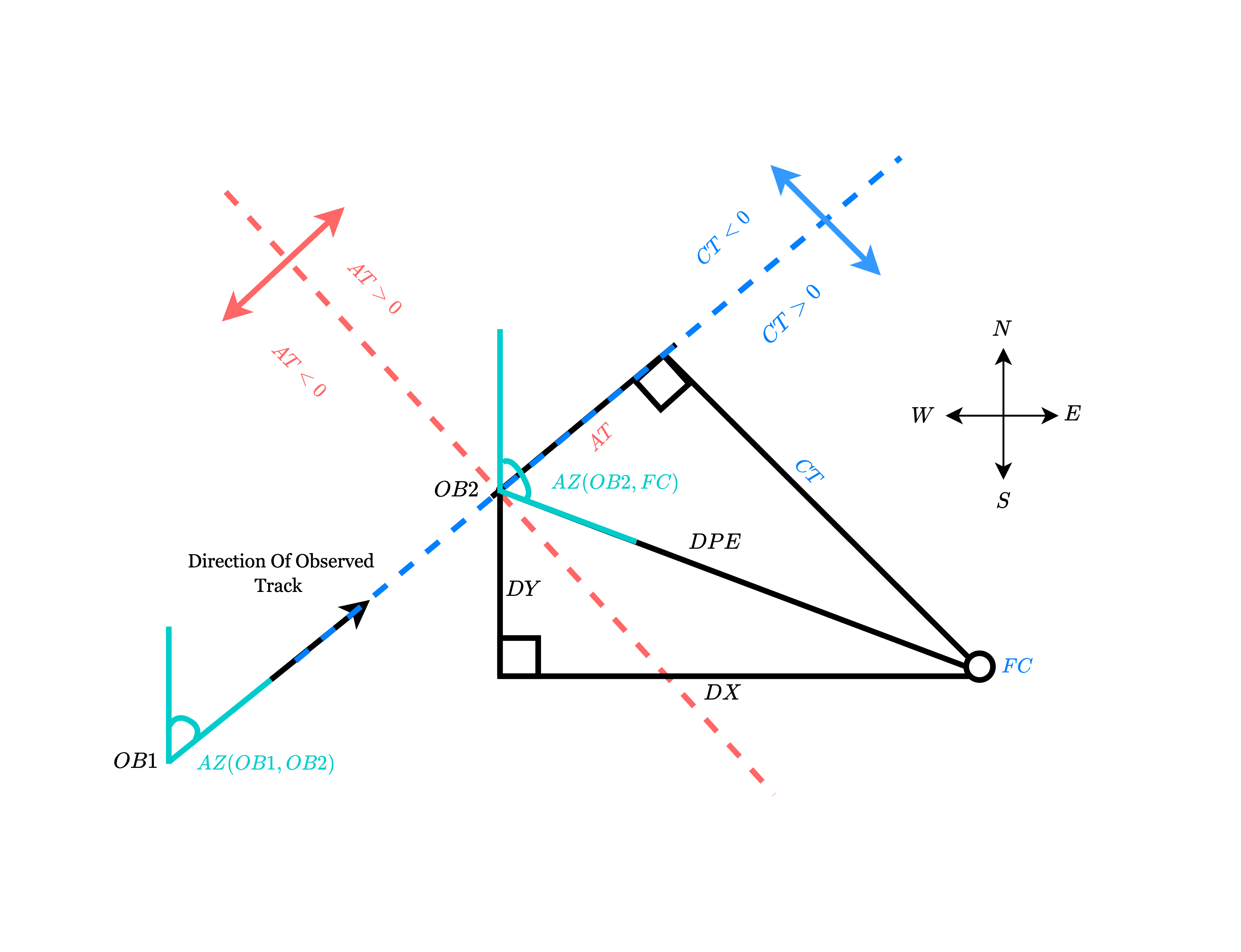}
    \caption{\textbf{Geometric schematic diagram of TC track forecast error decomposition.}
    This schematic diagram illustrates the calculation method for along-track error (AT) and cross-track error (CT) in TC track forecasting. $\mathrm{OB}_1$ and $\mathrm{OB}_2$ represent the TC center positions at two consecutive observation times, and FC represents the forecast position at the corresponding time. AZ($\mathrm{OB}_1$, $\mathrm{OB}_2$) is the azimuth angle of the observed track, defining the actual movement direction of the TC; AZ($\mathrm{OB}_2$, \text{FC}) is the azimuth angle from the latest observed position to the forecast position. DPE (Direct Position Error) represents the straight-line distance between the observed and forecast positions, which can be decomposed into horizontal component DX and vertical component DY. AT is defined as the projection error of the forecast position in the direction of the observed track: $\text{AT} > 0$ indicates the forecast position is ahead of the observed track, $\text{AT} < 0$ indicates the forecast position lags behind the observed track. CT is defined as the deviation of the forecast position perpendicular to the observed track direction: $\text{CT} > 0$ indicates the forecast position is on the right side of the observed track, $\text{CT} < 0$ indicates the forecast position is on the left side of the observed track. The red and blue dashed lines respectively mark the positive and negative regional divisions of AT and CT. The compass in the upper right corner indicates the geographic coordinate system (N-North, S-South, E-East, W-West). This decomposition method helps quantitatively analyze the systematic bias characteristics in TC track forecasting.}
    \label{model}    
\end{figure}

\FloatBarrier

\end{document}